\documentclass[11pt]{article}

\usepackage{fullpage}
\usepackage[ruled, vlined, linesnumbered, commentsnumbered]{algorithm2e}
\usepackage{graphicx,subcaption} 
\usepackage{amsfonts,amssymb,amsmath,amsthm,amsopn}	
\usepackage{hhline,booktabs,colortbl,multirow,tabularx,diagbox,threeparttable} 
\usepackage[listings,skins,breakable]{tcolorbox}

\usepackage{enumerate}
\usepackage{authblk}
\usepackage{footnote}
\usepackage{hyperref}
\usepackage{prettyref}
\usepackage{cite}
\usepackage{setspace}
\usepackage{color}
\usepackage{xcolor}  
\usepackage{lscape}
\usepackage{geometry}
\usepackage{relsize}

\usepackage{tikz}
\usepackage{pgfplots}
\usetikzlibrary{positioning,shapes,shadows,arrows,calc}
\tikzstyle{component}=[rectangle, draw=black, rounded corners, fill=blue!40, drop shadow, text centered, anchor=north, text=white, minimum height=1cm]
\tikzstyle{arrow}=[->, thick]

\pgfplotsset{compat=1.12}
\usetikzlibrary{intersections}
\usetikzlibrary{pgfplots.statistics}
\usepgfplotslibrary{fillbetween}

\SetKwFor{Parallel}{for each}{do in parallel}{end for}
\SetKw{Of}{of}

\tcbset{colback=white, arc=0mm, outer arc=0pt}

\def\hlinew#1{%
  \noalign{\ifnum0=`}\fi\hrule \@height #1 \futurelet
   \reserved@a\@xhline}
\makeatother

\definecolor{myblue}{RGB}{34,31,217}
\definecolor{mycyan}{gray}{.7}
\definecolor{Gray}{gray}{0.9}

\DeclareMathOperator*{\argmax}{argmax}

\let\oldnl\nl
\newcommand{\noln}{\renewcommand{\nl}{\let\nl\oldnl}}

\newcommand{\pref}{\prettyref}

\newrefformat{fig}{Fig.~\ref{#1}}
\newrefformat{tab}{Table~\ref{#1}}
\newrefformat{sec}{Section~\ref{#1}}
\newrefformat{app}{Appendix~\ref{#1}}
\newrefformat{alg}{Algorithm~\ref{#1}}
\newrefformat{property}{Property~\ref{#1}}
\newrefformat{theorem}{Theorem~\ref{#1}}
\newrefformat{corollary}{Corollary~\ref{#1}}
\newrefformat{proposition}{Proposition~\ref{#1}}
\newrefformat{def}{Definition~\ref{#1}}
\newrefformat{eq}{equation~(\ref{#1})}

\newcolumntype{R}{>{\raggedleft\arraybackslash}X}
\newcolumntype{L}{>{\raggedright\arraybackslash}X}
\newcolumntype{P}[1]{>{\raggedright\arraybackslash}p{#1}}

\newcommand{\abs}[1]{\left\lvert #1 \right\rvert}

\makeatletter
\def\BState{\State\hskip-\ALG@thistlm}
\makeatother

\SetCommentSty{mycommfont}

\title{\vspace{-1ex}\LARGE\textbf{Routing-Led Evolutionary Algorithm for Large-Scale Multi-Objective VNF Placement Problems}\footnote{This manuscript is submitted for potential publication. Reviewers can use this version in peer review.}}

\author[1]{\normalsize Peili Mao}
\author[2]{\normalsize Joseph Billingsley}
\author[3]{\normalsize Wang Miao}
\author[3]{\normalsize Geyong Min}
\author[3]{\normalsize Ke Li}
\affil[2]{\normalsize University of Electronic Science and Technology of China, Chengdu, UK}
\affil[2]{\normalsize G-Research, London, UK}
\affil[3]{\normalsize Department of Computer Science, University of Exeter, EX4 4QF, Exeter, UK}
\affil[$\ast$]{\normalsize Email: \texttt{k.li@exeter.ac.uk}}

\date{}

\begin{document}
\maketitle

\vspace{-3ex}
{\normalsize\textbf{Abstract: }	Modern data centers contain thousands of servers making them major consumers of electricity. To minimize their environmental impact, it is critical that we use their resources efficiently. In this paper we study how to discover the optimal placement of virtual network functions in large scale data centers. We propose a novel parallel metaheuristic, fast heuristic objective functions of the QoS and new memory efficient data structures for large networks. We further identify a simple, fast heuristic that can produce competitive solutions to very large problem instances. Using these new concepts, we are able to find high quality solutions for data centres with up to $64,000$ servers.


\section{Introduction}
Modern communication services have and continue to have an outsized impact on the way that society works \cite{OECD16}, travels \cite{ZhengMF15}, socializes \cite{Bargh04} and engages with politics \cite{Farrell12}, government \cite{ChadwickM13} and other individuals \cite{Castells14}. But this does not come without a cost. These services are also major consumers of electricity, with communications technology as a whole expected to be responsible for 21\% of the worlds total electricity usage by 2030 \cite{AndraeE15}.

If current trends continue, by 2030 data centers will likely be the largest single contributor of carbon emissions among communications infrastructure \cite{AndraeE15}. Over the previous decade, extensive work was conducted to reduce the energy consumed by overhead, such as heat management and energy provisioning, in data centers \cite{AvgerinouBC17}. Through these efforts, the energy total consumption of data centers remained constant in the United States and doubled in the European union \cite{DoddAGC20} from 2010 - 2020, despite a forecasted 10$\times$ increase in data center traffic. However, in recent years, progress in this direction has slowed, as improvements to overhead efficiency have reached the point of diminishing returns.

Recent research found that the efficiency of a data center could be improved by a further 10\% to 40\% by maximizing the utilization of existing computing equipment \cite{DoddAGC20,ShehabiARSSD16}. However, existing communication services make extensive use of specialized hardware which can only be utilized for a single task. Specifically, communication services are typically constructed using sequences of purpose built computing equipment known as middleboxes or physical network functions. Each middlebox can perform a single task with high efficiency, but being physical components, the number and location of these middleboxes must be specified well in advance of their usage. A recent alternative are Virtual Network Functions (VNFs). VNFs use software running on virtual machines to perform the same tasks as middleboxes. Using VNFs, multiple network functions can be provided by the same server, maximizing the utilization of hardware which results in less energy being needed to provide the same quality of service (QoS). In addition, the resources used by VNFs can be moved and scaled to meet traffic demands without over or under allocating resources.

Although VNFs allow increased utilization of servers, it remains an open problem how these VNFs should be assigned to servers. The VNF Placement Problem (VNFPP) is the task of automatically determining the placement of VNFs in a data center so as to provide high quality services with low energy consumption. The VNFPP presents many challenges. First, it is widely acknowledged that VNFPP is a NP hard problem~\cite{CohenLNR15,LuizelliCBG17,SangJGDY17} for which there are no known solutions that can find an exact solution in reasonable time. Second, whilst the QoS objectives and energy consumption are both well understood, they are challenging to model accurately and efficiently. Finally, due to the large numbers of servers and VNFs in a data center the total search space is very large whilst the feasible search space is relatively small \cite{BillingsleyLMMG22} which makes finding feasible solutions challenging, and high quality solutions even more so.

In prior work \cite{BillingsleyLMMG22}, we proposed a combined metaheuristic and model solution with a genotype-phenotype mapping that found good solutions to the VNFPP despite these challenges and which could solve $8\times$ larger problems than the state of the art. However, the proposed algorithm was only able to solve problems for one type of network topology. More recently, we proposed a modification of our original algorithm that allowed the algorithm to work on arbitrary network topologies but this approach had limited scalability \cite{BillingsleyLMMG20}. In this paper we build on our earlier work to make further improvements and resolve the scalability issues. In particular, this paper combines our generalized genotype-phenotype mapping with the following innovations:

\begin{enumerate}
    \item A new parallel metaheuristic that uses decomposition and local search to efficiently search the large solution space.
    \item Efficient objective functions that balance model accuracy against evaluation time.
    \item Memory efficient data structures that significantly reduce the memory consumption of the metaheuristic
\end{enumerate}

Using these innovations, our solution is the first to solve the VNFPP for $\sim$65,000 servers. This represents a further 4x improvement over our earlier solution, and a 16x improvement over the preceding state of the art.

The remainder of this paper is organized as follows. Section \ref{sec:literature_review} reviews the current work on the VNFPP. Section \ref{sec:problem_formulation} provides a formal definition of the VNFPP. Section \ref{sec:optimisation} describes the parallel metaheuristic and operators considered in this work including the memory efficient data structures. Section \ref{sec:practical_objective_functions} discusses common objective functions for the VNFPP. In Section \ref{sec:evaluation} we test the effectiveness of each component of the algorithm and evaluate our solution on very large problem instances. Finally, Section \ref{sec:conclusion} concludes this paper and outlines some potential future directions.

\section{Related Works}
\label{sec:literature_review}

This section provides a pragmatic overview of some important developments on the VNFPP. Whilst there is a wealth of work on the problem, most solutions have limitations that make them impractical for real world application. Existing solutions tend to fall into two groups:
\begin{itemize}
    \item Work that proposes solutions for realistic versions of the problem but which cannot scale to large problems.
    \item Work that finds a scalable solution to a simplified problem that may not be useful in practice.
\end{itemize}

\subsection{Realistic but Not Scalable}
Several works that use realistic models of the data center in the VNFPP also use exact methods to optimise them. Exact methods guarantee optimal solutions but also have an exponential worst case time complexity for NP-Hard problems such as the VNFPP~\cite{Landa-Silva13} and as a result, can only solve problems with tens of servers, making them impractical for real world problems. Further, exact methods typically require a linear objective function whilst real measurements of the QoS show it to be non-linear \cite{IntelDPDK, IntelPPP}. In order to use a realistic model with exact methods, researchers commonly use piecewise linearization to linearize the measures of QoS. For example, Addis et al.~\cite{AddisBBS15} used linear programming and piecewise linearization to solve two VNFPPs: one where waiting time is modelled as a convex piece-wise linear function of the sum arrival rate, and another where the latency is a constant when it is below a certain threshold. This work considered up to 15 servers and solved two objectives, by first minimizing the max link utilization and then minimizing the number of cores used. Gao et al.~\cite{GaoABS18} extended this work to allow one VNF per service to alter the traffic rate leaving the VNF. In \cite{BaumgartnerRB15}, Baumgartner et al. place a small number of VNFs across a geographically distributed network. They model the processing, transmission and queueing delays and linearize them with piecewise linearization. Oljira et al.~\cite{OljiraGTB17} used the same techniques as \cite{BaumgartnerRB15} for modeling and optimization and additionally considered the virtualization overhead when calculating the latency at each VNF. Both of these works considered a problem where there were only 28 locations a VNFs could be placed. Similarly, Jemaa et al.~\cite{JemaaPP16} only allowed VNFs to be placed in two locations: a resource constrained cloudlet data center near the user and an unconstrained cloud data center. VNFs were assigned to a data center to optimise for a combination of latency and cloudlet and cloud utilization. Agarwal et al.~\cite{AgarwalMCD18} also use exact methods to solve a small problem instance with 3 servers. They aim to minimize the ratio of the actual latency to the max latency for each service. Marotta et al.~\cite{MarottaZDK17} use a series of heuristics and exact methods to solve problem instances with 10s of servers. They first assign VNFs to servers to minimize energy and inter-server traffic. Then a heuristic makes the placement robust to changes in arrival rate and finally exact methods are used to find routes between VNFs to form services that minimize the energy consumption whilst meeting latency constraints.

\subsection{Scalable but Unrealistic}
The other class of impractical solutions use heuristic or metaheuristic algorithms to provide scalable solutions but that solve oversimplified versions of the VNFPP. This research can be subdivided into solutions that use heuristic objective functions, solutions which use unrealistic models of the QoS, and solutions that do not utilise a model at all. 

\subsubsection{Heuristic Model}
Several works do not consider the QoS or energy consumption and instead optimise for simpler models that are easier to construct and solve. These works do not consider whether these heuristics are suitable metrics compared to an accurate model and hence may not actually be optimizing for the intended objectives and hence may produce subpar solutions. For example, we have previously shown that if common heuristic models are used for the VNFPP the solutions are significantly less diverse than those found by accurate models \cite{BillingsleyLMMG22}.

Kuo et al. \cite{KuoLLT18} aim to maximize the number of placed services by efficiently reusing network functions. They first preprocess each service with ILP to find a structure that allows for the most VNF reuse. Then they use dynamic programming to place VNFs, aiming to minimize the consumed resources. Guo et al. \cite{GuoWLQA0Y20} also reuse VNFs but aim to minimize bandwidth and placement costs. They first preprocess the topology to find the most influential nodes according to the Katz centrality. Shareable VNFs can only be placed on these nodes increasing the chance they are reused. Finally they place VNFs using a markov decision process with lower costs for reusing VNFs. Qi et al. \cite{QiSW19} use greedy search to minimize the energy and bandwidth used by the solution. To increase the speed of the search they only considered nodes that were within some number of hops from the current node. Rankothge et al. \cite{RankothgeLRL17} use a genetic algorithm and custom operators to place VNFs, aiming to minimize link utilization and the number of servers used whilst maximizing the number of links used.

\begin{figure*}[t]
    \begin{minipage}{0.425\columnwidth}
        \centering
        \includegraphics[width=1\textwidth]{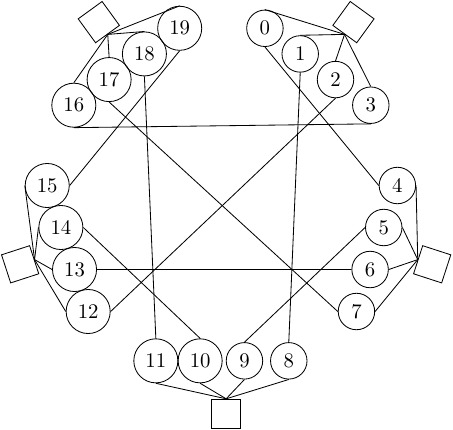}
        \subcaption{Dcell$_1$ with 4 port switches \cite{GuoWTSZL08}}
    \end{minipage}\hfill
    \begin{minipage}{.525\columnwidth}
        \includegraphics[width=\textwidth]{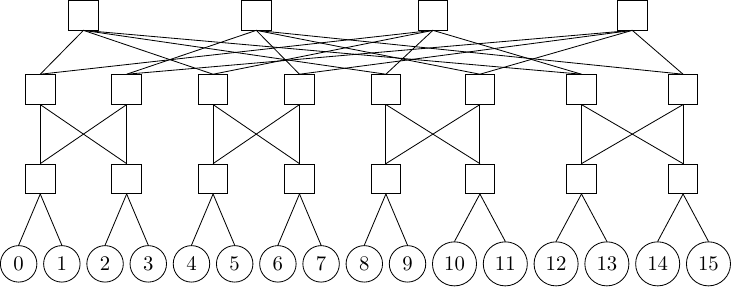}
        \subcaption{Fat Tree with 4 port switches \cite{Al-FaresLV08}}

        \vspace{2.5em}

        \includegraphics[width=\textwidth]{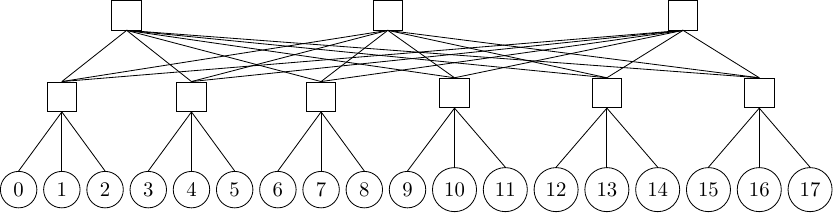}
        \subcaption{Leaf-Spine with 6 port switches \cite{Cisco19}}
    \end{minipage}\hfill
    \begin{minipage}{.05\columnwidth}
        \includegraphics[width=\textwidth]{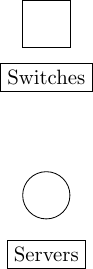}
    \end{minipage}

    \vspace{1em}
    \caption{Common data center topologies}
    \label{fig:topologies}

	\vspace{1em}
\end{figure*}

\subsubsection{Unrealistic Model}
Other works do consider the QoS but do so in a way that is inconsistent with real world evidence. Most commonly, these works assume that the waiting time at a network component is constant however in practice the waiting time depends on the utilization of the component \cite{IntelDPDK, IntelPPP, OljiraGTB17}.

Many works assume there are constant waiting times at all components in the network. Often, these works use greedy heuristics to solve the simplified problem. For example, Alameddine et al. \cite{AlameddineQA17} proposed greedy, random and tabu heuristics to place VNFs such that tasks are completed before certain deadlines. Vizarreta et al. \cite{VizarretaCMMK17} also assumed constant waiting times but additionally fixed the starting and ending components for each service. They first found the lowest cost route between the starting and ending components that satisfied latency and robustness constraints and then modified the path until it could accommodate each VNF of the service. Qu et al. \cite{QuASK17} used the same assumptions and a similar technique. Their algorithm first attempts to place VNFs on the shortest path between the starting and ending components of each service. If the path cannot accommodate all VNFs then nearby components are also considered.

Other non-greedy heuristics also presuppose constant waiting times. These include Hawilo et al. \cite{HawiloJS19}, who proposed a heuristic named BACON that places latency sensitive VNFs on components with high betweeness centralities and low latency neighbors. Manias et al. \cite{ManiasJHSHLB19} later extended this work, using a decision tree to learn the outputs of BACON. More recently they extended this work to use particle swarm optimization to select the model hyperparameters \cite{ManiasHJS20}. Roy et al. \cite{RoyTSARH20} use ant colony optimization to minimize the overall latency where the ants heuristically favoured nodes with low constant waiting times. Bari et al. \cite{BariCAB15} used dynamic programming to place each service separately aiming to minimize energy and placement costs whilst considering latency constraints.

Using the same constant waiting time assumption, some researchers have combined linear programming with other methods to reduce the number of possibilities the solver must consider. Alleg et al. \cite{AllegKMA17} consider a problem with branching services and used a heuristic to match servers and VNFs by their degree in the network topology and service chain respectively. The solver is then only allowed to place VNFs on those nodes in order to meet latency constraints. Luizelli et al. \cite{LuizelliCBG17} used variable neighborhood search (VNS) to selects subsets of VNFs that can be changed whilst the others variables remain fixed. The new problem is then solved whilst meeting latency constraints using linear programming. A new subset of VNFs are selected for the next iteration and the process is repeated. Pei et al. \cite{PeiHXLWW20} used linear programming to find the exact solution to a set of problems to form a training set and then used supervised learning to learn a model that can place VNFs.

\subsubsection{Ignores Model}
Some researchers recognize the relevance of accurate QoS models but only use it to evaluate the solutions discovered by a heuristic. This provides additional information to the decision maker but does not change the quality of the solutions. Chua et al. \cite{ChuaWZSH16} place VNFs of each service in the first server that can accommodate it, but restrict the maximum capacity of each server until all available capacity has been used. They then use an queueing model to inform the end user of the expected latency. Similarly, Zhang et al. \cite{ZhangXLLGW17} propose a best fit decreasing method of placing VNFs and use an queueing model to evaluate the solution.

\subsection{Realistic and Scalable}
Finally, there are some works that meet both of the main criteria but which may be improved upon. Some works use more accurate but still imperfect models with known issues. For example, several works disregard packet loss when constructing their model. Whilst this greatly simplifies the model construction it also misrepresents the problem as packet loss impacts on service latency and energy consumption \cite{BillingsleyLMMG22}. Gouareb \cite{GouarebFA18} et al. use this type of model to calculate the latency, and present linear programming and heuristic solutions that minimize the latency. The heuristic greedily assigns VNFs with the largest requirements to the server with the most capacity and then scales VNFs horizontally (more instances) or vertically (more resources per VNF) to meet demand. Leivadeas et al. \cite{LeivadeasFLIK18} use the same form of model with Tabu search to jointly minimize energy and latency. Qu et al. \cite{QuZYSLR20} use the same model to minimize cost whilst considering latency constraints. They propose a heuristic that attempts to reallocate resources on overloaded nodes to services which are violating constraints and if necessary to migrate services so as to avoid overloaded nodes.

In our earlier work on this problem \cite{BillingsleyLMMG20}, we proposed a routing-led multi-objective genetic algorithm that used an accurate queueing model of a data center. This earlier work could scale to problems with up to $\sim$16,000 servers. On larger problems, the time and memory requirements of the algorithm are impractical.




\section{Problem Formulation}
\label{sec:problem_formulation}

In this section, we formally define the VNFPP and its constraints. First, we describe a high level overview of the problem and then provide a formal problem statement.

The aim of the VNFPP is to find a solution that maximizes the QoS of each service and minimizes the total energy consumption. A service is formed by directing traffic through VNFs in a specific order where each VNF of the service is placed in the data center. Each VNF requires a certain amount of resources, and can be assigned to any server that has sufficient resources available. The data center consists of a large number of servers with finite computational resources. These servers can communicate using the network topology, a set of switches that interconnect all servers (see Fig. \ref{fig:topologies}), to provide services that require more resources than one server can provide. Hence a service can be constructed by assigning VNFs to servers with sufficient capacity and describing paths that connect them using the network topology.









We can now formally define the VNFPP. First, we define some core terminology and then we define the objectives and constraints of the problem. $\mathcal{S}$ is the set of services that must be placed and $\mathcal{V}$ is the set of all VNFs. A service is a sequence of VNFs: $s \in S$, $s = \{s_1, s_2, ..., s_n\}$. The network topology is represented as a graph $\mathcal{G}=(\mathcal{C},\mathcal{L})$, where $\mathcal{C}$ denotes the set of data center components and $\mathcal{L}$ denotes the set of links connecting those components. A route is a sequence of data center components where $R^s$ is the set of paths for service $s$, $R_{i}^s$ is the $i$th path of the service and $R_{i,j}^s$ is the $j$th component of the path. Finally, $\abs{\cdot}$ gives the cardinality of a set or a sequence. 

The resource and sequencing constraints of the problem can be formalized as follows:

\begin{enumerate}
	\item Sequential components in a route must be connected by an edge:
		\begin{equation}
			(R_{i,j}^s, R_{i,j+1}^s) \in \mathcal{L}
		\end{equation}
    \item The sum resources required by the VNFs assigned to a server cannot exceed the maximum capacity of the server:
		\begin{equation}
			\sum_{v \in A_{c_\mathsf{sr}}} C_v < C^{\mathsf{sr}}
        \end{equation}
		where $A_{c_\mathsf{sr}}$ is the set of VNFs $v$ assigned to server $c_\mathsf{sr}$, $C_v$ is the resources required by the VNF $v$ and $C^{\mathsf{sr}}$ is the total resources available on a server.
	\item All VNFs must appear in the route and in the order defined by the service:
		\begin{align}
			& \pi^{R^s}_{s_i} \neq \emptyset \\
			& \pi^{R_i}_{s_i} < \pi^{R_i}_{s_{i+1}} 
		\end{align}
		where $\pi^{R_i}_{S_i}$ is the index of the VNF $s_i$ in route $R_i$.
\end{enumerate}

We consider a three-objective VNFPP including two critical metrics of the QoS, latency and packet loss, and the total energy consumption. As service latency, packet loss and energy consumption can conflict \cite{BillingsleyLMMG22} we formulate the VNFPP as a multi-objective optimization problem. Further, as the number of services in a data center could number in the thousands it is not practical to treat each service quality metric as a separate objective due to the curse of dimensionality. Instead we aim to minimize the aggregate latency and packet loss and the total energy consumption of the data center. Formally, these objectives are defined as:

\begin{itemize}
	\item The \textit{total energy consumption} $\mathbf{E}$.
    \item The \textit{mean latency} of the services $\mathbf{L}$:
        \begin{equation}
            \mathbf{L}=\sum_{s \in S} W_{s}/|S|,
        \end{equation}
        where $W_s$ is the expected latency of the service $s\in S$. 
    \item The \textit{mean packet loss} of the services $\mathbf{P}$:
        \begin{equation}
            \mathbf{P}=\sum_{s \in S} \mathbb{P}^d_s / |S|,
        \end{equation}
        where $\mathbb{P}^d_s$ is the packet loss probability of the service $s$.
\end{itemize}

The exact energy consumption and latency and packet loss of each service for a solution can be found using tools such as discrete event simulation \cite{Pongor93}. However, VNFPP solvers rarely use exact measurements as procuring them can be time consuming. Commonly, accurate models or heuristics of the objective functions are used during the optimization process. Alternative objective functions are discussed in detail in \pref{sec:practical_objective_functions}.

\section{Improved EMO Algorithm for the VNFPP}
\label{sec:optimisation}
In this section we propose a new algorithm for the VNFPP that can find high quality solutions to large scale VNFPPs. There are three key components of our algorithm: 
\begin{enumerate}
    \item A novel parallel metaheuristic that efficiently searches the VNFPP solution space.
    \item A genotype-phenotype solution representation that incorporates domain knowledge to improve expected solution quality.
    \item A novel initialization operator that ensures a diverse initial pool of feasible solutions.
\end{enumerate}
\pref{fig:overview} provides an overview of how these components work together. In the first stage, \textit{preprocessing}, the algorithm decomposes the multi-objective problem into subproblems and generates data structures that aid in efficient optimization. In the second stage, \textit{optimization}, the algorithm finds solutions to each subproblem in parallel.

The remainder of this section discusses each component of the algorithm in detail. First we discuss the core optimization algorithm and the ways it improves upon from existing parallel metaheuristics in \pref{sec:core_algorithm}. Next we describe the tailored solution representation we use to aid the search process in \pref{sec:gp_mapping}. Finally, in \pref{sec:operators} we discuss the initialization and local search operators used in this work. 

\begin{figure}
    \includegraphics{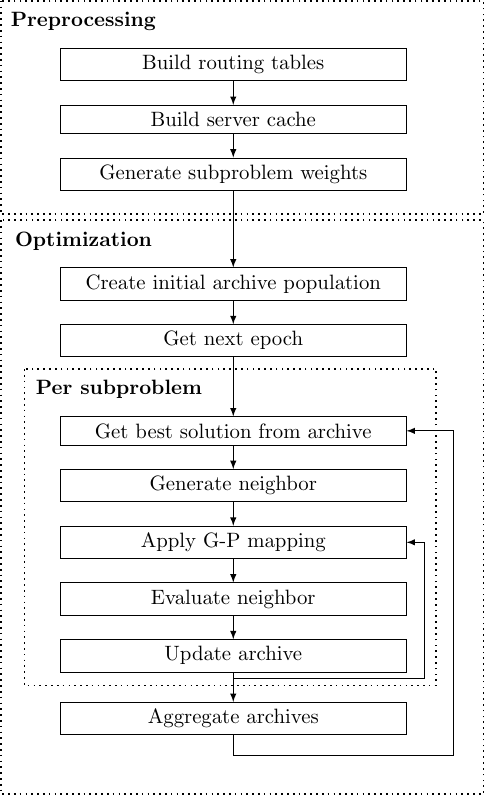}
    \caption{A high level overview of our proposed algorithm.}
    \label{fig:overview}
\end{figure}

\subsection{Novel Parallel Metaheuristic}
\label{sec:core_algorithm}

\begin{algorithm}[t!]
	
	\KwData{Subproblems per epoch $N^s$, Number of processes $L$, Weight vectors $W = \{W^1, ..., W^N\}$, Max evaluations $T$, Number of objectives $M$}

    $N^E \gets \lceil \abs{W} / N^s \rceil$ \tcp{Number of epochs}
    $T^W \gets (T - \abs{A}) / \abs{W}$ \tcp{Evaluations per weight} 

    \tcc{Evaluate initial solutions}
    $A' \gets \textsc{Initialize}$
	\For{$s \in A'$} {
        $s_{obj} \gets \textsc{Evaluate} (s)$
	}
    $A \gets$ Non-dominated pop. of $A'$

    \For{$e \gets 0$ \KwTo $N^E$} {

        $i_f \gets e \cdot N^s$ \tcp{Index of first weight}
        $i_l \gets \min(i_f + N^s, \abs{W})$ \tcp{Index of last weight}

        \Parallel{$\ell$ in $\{W_{i_f}, ... W_{i_l}\}$} {
            $A_\ell \gets \emptyset$ \tcp{Process archive}

            \tcp{Best solution found for weight}
            $s \gets \underset{x \in A}{\argmax}\ g \left( x\ |\ W_\ell \right)$

            $A_\ell \gets A_\ell \cup s$

            $t \gets 0$ \tcp{Evaluation counter}

            \While{$t < T^W$} {
                \tcc{Get neighbouring solution}
                $s^n \gets \textsc{GenNeighbour}(s)$

                $s^n_{obj} \gets \textsc{Evaluate} (s^n)$

                $t \gets t + 1$

                \tcc{Update process archive}
                $A_\ell \gets$ Non-dominated pop. of $A_\ell \cup s^n$

                \tcc{Accept improving solutions}
                \If{$g\left(s^n | W_\ell \right) < g \left(s | W_\ell \right) $}{
                    $s \gets s^n$
                }
            }
        }

        $A \gets$ Non-dominated pop. of $(\bigcup_\ell^L A_\ell)$
    }

	\caption{Our proposed decomposition based parallel multi-objective local search algorithm.}
	\label{alg:simple_pplsd}
\end{algorithm}

It is well known that the Pareto front of a problem can be approximated by solving a diverse set of scalar optimization subproblems. For example, given the multi-objective optimization problem:
\begin{equation}
    minimize\ F(x) = (f_1(x), ..., f_m(x))^\intercal
    \label{eq:mop}
\end{equation}
where $x \in \Omega$ and $\Omega$ is the decision space, $F: \Omega \implies R^m$ consists of $m$ real valued objective functions. Provided the objectives in \pref{eq:mop} contradict each other, no point in $\Omega$ minimizes all objectives simultaneously. However, given a weight vector $\mathbf{\lambda} = \{\lambda^1, ... \lambda^n\}$ (where $\lambda_i \geq 0 \forall i = 1,\cdots,m$ and $\sum_{i=1}^m \lambda_i = 1$) and a utopian point ((i.e. $z_i^* = min \left\{f_i(x) | x \in \Omega \right\}$)),(i.e. $z_i^* = min \left\{f_i(x) | x \in \Omega \right\}$) the optimal solution to the Tchebycheff \textit{scalar} optimization problem:
\begin{equation}
    minimize\ g^{tc}(x | \lambda, z^*) = \max_{1 \leq i \leq m} \left\{\lambda_i \abs{f_i(x) - z_i^*} \right\},
    \label{eq:tchebycheff}
\end{equation}
is a Pareto optimal solution to \pref{eq:mop}. Hence, by varying the weight vector $\lambda$ we can generate a set of subproblems for which the solution is a set of Pareto optimal solutions. 

In order to solve large multi-objective problems, we can decompose the problem into a set of scalar subproblems and solve them in parallel. However, there exists a clear tradeoff between communicating information on solutions between subproblems and the work that can be performed in parallel. Decomposition based algorithms such as MOEA/D share solutions between subproblems to facilitate faster convergence but at the cost of limited capacity for parallelization. Alternatively, existing parallel decomposition algorithms such as PPLS/D \cite{ShiZS20}, solve each subproblem in isolation, preventing any inter-problem communication but maximizing the parallelized work.

We propose a new algorithm which utilizes a common archive to allow the degree of communication to be determined by a parameter. The pseudo-code of our algorithm is listed in \pref{alg:simple_pplsd}. The algorithm can be split into four key stages:
\begin{enumerate}
    \item We decompose the problem into $L$ subproblems and group them into `epochs' where each epoch contains $N$ subproblems (lines 1 - 2).
    \item We generate an initial set of solutions (lines 2-3, \pref{sec:initialization}) and add the non-dominated ones to the \textit{total archive} (lines 4 - 5).
    \item For each subproblem in each epoch, we select the best solution from the total archive as a trial solution (line 14) and then use local search to search for a solution to the subproblem (line 16-25). During this search process, non-dominated solutions are added to the \textit{subproblem archive} (line 22).
    \item The subproblem archives are merged into the total archive. The total archive is used in future epochs to select better trial solutions (line 26). By varying the number of subproblems in an epoch, the frequency of communication can be altered.
\end{enumerate}

Our proposed algorithm can be applied to any multi-objective optimization problem however better results for the VNFPP can be achieved with further work. There are three key factors to consider: the choice of scalar objective function, the number of epochs and constraint handling. 

\subsubsection{Scalar objective function}
It is well known that decomposition based multi-objective evolutionary algorithms are less effective on problems with disparately scaled objective functions \cite{ZhangL07}. On large VNFPP problem instances, the disparity between objectives can be very high. A common way to resolve this issue is to normalize each objective, i.e.:
\begin{equation}
    \bar{f_i} = \frac{f_i - z^*_i}{z_i^{nad} - z_i^*}
\end{equation}
where $z_i^{nad} = \left(z_1^{nad},...,z_m^{nad}\right)^\intercal$ is the nadir point, (i.e. $z_i^{nad} = max \left\{f_i(x) | x \in \Omega \right\}$). This causes the range of each objective to be in $[0,1]$.

It is difficult and unnecessary to compute $z^{nad}$ and $z^*$ in advance. In our implementation, we approximate $z^{nad}$ and $z^*$ before each epoch using the highest and lowest objective values found in the total non-dominated archive. For this work we use the normalized Tchebycheff scalar objective function:
\begin{equation}
    \mathrm{minimize}\max_{1 \leq i \leq m} \left\{\lambda_i \abs{\frac{f_i - z_i}{z_i^{nad} - z_i^*}} \right\}.
\end{equation}

\subsubsection{Number of epochs}
The number of epochs determines how frequently information can be communicated between subproblems but also determines the amount of work that can be performed in parallel. Since fast execution time is a priority in this work, we use the maximum number of processes available in each epoch to maximize the amount of work conducted in parallel.
 
\subsubsection{Constraint handling}
Finally, whilst \pref{alg:simple_pplsd} does not explicitly provide methods for accommodating constrained problems such as the VNFPP, it is trivial to introduce this requirement by modifying which solutions can be accepted into the archive. For this problem, we extended dominance such that feasible solutions dominate infeasible solutions. This is sufficient since the initialization operator typically generates feasible solutions that populate the archive.

\subsection{Genotype-Phenotype Mapping}
\label{sec:gp_mapping}

\begin{figure}
    \includegraphics{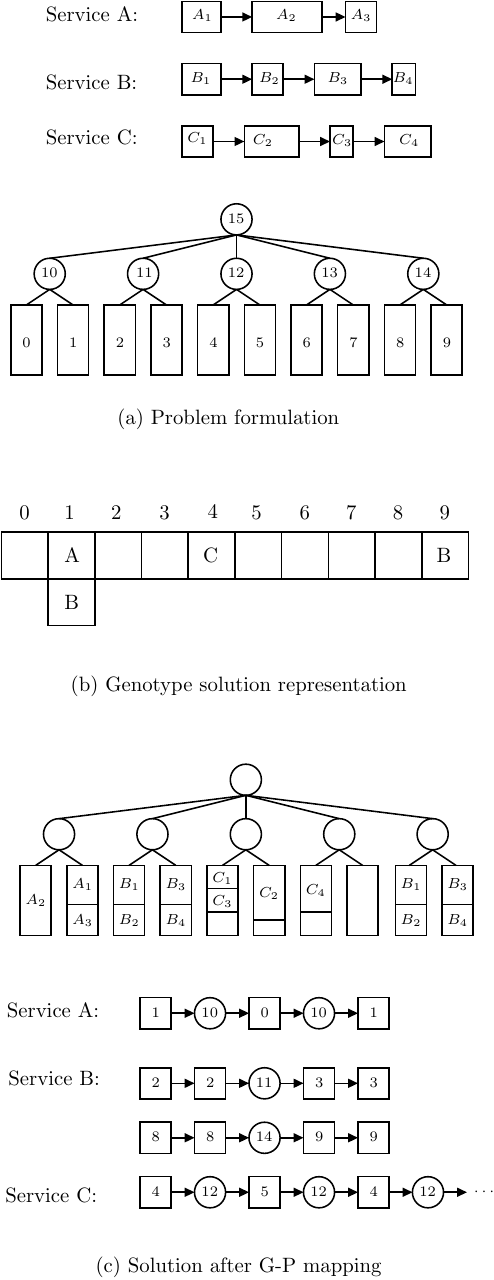}
    \caption{An example of the components of the solution representation.}
    \label{fig:mapping}
\end{figure}


One of the key challenges when designing an EA for real-world problems is selecting the right solution representation. For complex problems such as the VNFPP, it can be beneficial to introduce domain knowledge into the solution representation to improve the expected solution quality. One way this can be performed is using a \textit{genotype-phenotype} (G-P) solution representation. A G-P solution representation defines a `genotype' solution representation ($\Omega'$) and a G-P mapping function to convert it into the true solution space, f:$\Omega' \implies \Omega$. Domain knowledge can be introduced into the genotype and in the G-P mapping function.

In this work we use a G-P solution representation to improve the probability of generating a feasible solution and to minimize the distance between sequential VNFs. The genotype is a list of servers where each server can contain any number of \textit{service instances} - an indicator that a service should start from that location. \pref{fig:mapping} illustrates how the G-P mapping is performed. The mapping algorithm iterates over each service instance, and maps each VNF of the service to the nearest server that can accommodate it (\pref{fig:mapping} iii). Once all VNFs have been placed, the mapping then finds the set of shortest paths between the VNFs of the service and assigns them to the solution (\pref{fig:mapping} iv).

\begin{algorithm}[t!]
	

	\KwData{Initial server $s$, cache size $n_{max}$}
	\KwResult{Server distance cache $C_s$}

    $C_s \gets \emptyset$
    
    $Q_c \gets {s}$ \tcp{Current horizon}

    $Q_n \gets \emptyset$ \tcp{Next horizon}

    $E \gets \emptyset$ \tcp{Explored nodes}

    \While{$Q_c \neq \emptyset$} {
        $u \gets \text{Get random element of } Q_c$

        $Q_c \gets Q_c \ u$
        
        \If{$u \text{ is a server}$}{
            $C_s \gets C_s \cup u$

            \If{$\abs{C_s} == n_{max}$} {
                \Return{$C_s$}
            }
        }

        \ForEach{$\text{neighbour } v $ \Of $u$}{
            \If{$v \notin E$}{
                $Q_n \gets Q_n \cup v$

                $E \gets E \cup v$
            }
        }

        \If{$Q_c = \emptyset$}{
            $Q_c \gets Q_n$

            $Q_n \gets \emptyset$
        }
    }

	\caption{A non-deterministic breadth first search algorithm.}
	\label{alg:nd_bfs}
\end{algorithm}

\begin{figure*}[t]
    \includegraphics[width=\textwidth]{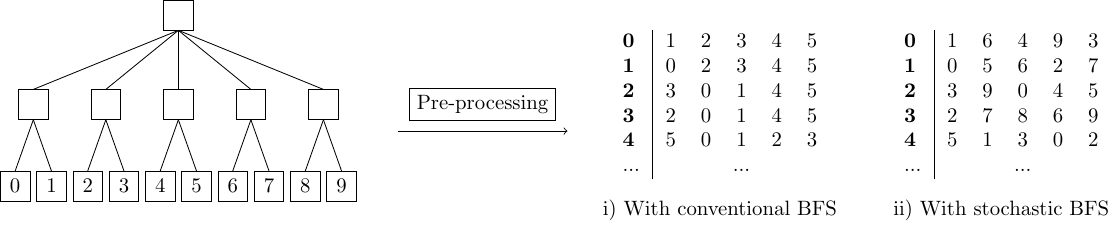}

    \caption{The results of generating the server cache with a conventional BFS vs a stochastic BFS}
    \label{fig:server_gen}
\end{figure*}

The G-P mapping function uses two preconstructed data structures to run efficiently: a set of routing tables for each network component and a set of distance tables for each server.
\begin{itemize}
    \item \textit{Distance tables} are used to find the next nearest server with sufficient capacity to place a VNF. Each distance tables simply list the other servers in the data center in increasing order of distance.
    \item \textit{Forwarding tables} are used to find the shortest routes between two servers. Each node in the graph has a forwarding table which can be consulted to find the next steps towards a server.
\end{itemize}

Naive implementations of these constructs do not scale well to large problems since the memory required scales with the number of servers and switches \cite{BillingsleyLMMG20}. In this work we propose compressed versions of both data structures for use on large problems.

\subsubsection{Distance table compression}
The distance tables can be constructed using breadth first search (BFS) which will encounter every other server in the data center in increasing order of distance. In a naive implementation, each distance table will store a reference to every other server, for a $O(n^2)$ total memory complexity across all distance tables. However, since the mapping procedure will only consult the table until it finds a suitable server, each table only needs to contain as many servers as are likely to be accessed.

The compressed version of the table can be crafted by simply halting the search early. However, if the distance tables do not contain all servers, a `cache miss' can occur if the mapping procedure does not find a suitable server in the distance table. As illustrated in \pref{fig:server_gen}, deterministic methods of identifying nearby servers such as BFS can accentuate this issue as a small number of servers will be overrepresented on average. 

We resolve this issue using a non-deterministic BFS (\pref{alg:nd_bfs}). The non-deterministic BFS modifies the well known BFS by exploring the neighbors of the current frontier in a random order. Since nodes that are encountered earlier are still explored earlier, the nearest nodes to a server are still discovered first. However, all servers which are the same distance away are equally likely to appear in the distance table.


\subsubsection{Forwarding table compression}
We create the forwarding tables using a modified version of the initialization process of the equal-cost multi-path (ECMP) routing protocol. In the first step, a server broadcasts a message listing its ID and the distance the message has travelled so far. Upon receiving a message, if the component has received a message for that server with a lower distance it will discard it. Otherwise it will record the origin of the message as one of the next steps to the server, increment the distance and rebroadcast the message. Once all network components have received a message from all servers, the algorithm terminates.

Due to the large number of servers and switches in a data center, it is infeasible to store uncompressed forwarding tables in memory. As data centers must support high numbers of servers each port must facilitate access to many servers. Hence, we can often aggregate several rules which list the same hops together into a single rule to save memory. 

In this work, we aggregate rules that pertain to servers with sequential IDs. Each rule specifies the range of servers it encompasses and the next hop for which any server in that range should take. We apply these compression rules for each forwarding table after each server broadcast occurs. 

The memory saving from this approach depends on the network topology. In the worst case, this approach has the same worst case memory complexity as the naive implementation. In practice however, we find it results in significant memory savings (see Section \ref{sec:evaluation}). 


\subsection{Operators}
\label{sec:operators}
Finally we discuss the design of the operators used in this work. Our proposed metaheuristic requires two operators: an initialization operator that provides seed solutions to the subproblems and a neighbor generation operator that generates a solution in the neighborhood of a given solution and is an integral part of the local search procedure.

\subsubsection{Initialization}
\label{sec:initialization}
Good initial solutions can significantly improve the final quality of the solutions to single objective problems \cite{VlasicDJ19,RamseyG93,OsabaDCOL14} such as the subproblems in our algorithm. Since the weights are uniformly distributed over the objective space, it is important that the archive contains a diverse population. We designed an initialization operator that produces diverse initial solutions for the first archive in order to seed the search process with good initial solutions.


To ensure the population is diverse, we determine the minimum and maximum number of service instances the data center can accommodate and then generate solutions uniformly over this space. A feasible solution has at least one instance of each VNF, hence the minimum capacity that will be used is the sum of the size of all VNFs,

\begin{equation}
    M^\textsf{min} = \sum_{v\in \mathcal{V}} C_v
\end{equation}

\noindent Similarly, a solution cannot place more VNFs than there is capacity for, hence the maximum capacity is the total capacity of the data center,

\begin{equation}
    MD^\textsf{max} = n \cdot C^s
\end{equation}

\noindent Next, we determine how much capacity each solution should be allocated. For a population with $n$ individuals, we permit the $i$th individual to use at most $i / n$\% of the total capacity. The number of instances of each service is determined by:
\begin{equation}
    N_v^i = \left(M^\textsf{min}\ /\ M^\textsf{max} - 1 \right) \cdot \frac{i}{n} + 1.
\end{equation}
where the whole number part determines the guaranteed number of each instance whilst the fractional part determines the probability an additional instance is placed. This ensures that expected number of service instances meets the target capacity.

Once the number of instances of each service have been calculated, the service instances are distributed uniformly at random over the initial solution.

\subsubsection{Neighbor Generation}
\label{sec:local_search}
The local search procedure aims to find improving solutions by exploring the immediate neighborhood of a solution. The neighborhood of the VNFPP contains changes in the number and placement of service instances. To explore this space, our neighbor generation function has an equal probability to add or remove a service instance or to move a service instances to a new server.

\section{Practical Objective Functions for Large Scale VNFPPs}
\label{sec:practical_objective_functions}
Since the objective function will be used many times per run, a suitable objective function is critical. Two types of objective function are often used on the VNFPP: accurate models and heuristic models. Accurate models can find accurate estimates of the objective functions by modelling how a typical packet moves through the data center. Heuristic models instead use fast heuristics which correlate with the objective functions, e.g. a common heuristic for latency is the average path length. Heuristic models typically evaluate solutions faster than accurate models but existing proposals are ineffective on multi-objective VNFPPs \cite{BillingsleyLMMG22}. In this work we evaluate models of both types in order to better understand the tradeoffs between model efficiency and effectiveness.

\subsection{Accurate models}
Accurate models of the VNFPP typically use queueing theory to model the average flow of packets through each component in the data center. A queueing model (QM) will calculate the expected number of packets arriving at each component in the data center based on some assumptions. Next, the component waiting time ($W_c$), packet drop probability ($\mathbb{P}^d_c$) and utilization ($U_c$) can be determined. These metrics can then be used to calculate the average QoS for each service. 

Specifically, the average latency is the sum of the expected waiting time, calculated by:
\begin{equation}
	W_s=\sum_{i=1}^{|R^s|} W_{R^s_i} \cdot P_{R^s_i},
    \label{eq:wt_service}
\end{equation}
where $W_{R^s_i}$ is the average latency for the path $R^s_i$. It is calculated as the sum of the waiting time at each network component:
\begin{equation}
	W_{R^s_i} = \sum_{c\in R^s_i} W_c.
    \label{eq:wt_path}
\end{equation}
Similarly, the packet loss is the probability a packet does not complete the service which is calculated with the expected packet loss over each path: 
\begin{equation}
    \mathbb{P}^d_s=\sum_{i=1}^{|R^s|} \mathbb{P}^d_{R^s_i}\cdot \mathbb{P}_{R^s_i},
	\label{eq:pl_service}
\end{equation}
where $\mathbb{P}^d_{R^s_i}$ is the probability that a packet is dropped on the path $R^s_i$ which is calculated as:
\begin{equation}
	\mathbb{P}^d_{R^s_i}=1-\prod_{c\in R^s_i}\left(1-\mathbb{P}^d_c\right).
	\label{eq:pl_path}
\end{equation}

Similarly, the expected energy consumption is the sum energy consumption of each component. In this work, we use a three state model of energy consumption where a component can be either \texttt{off}, \texttt{busy} or \texttt{idle}. A component is off if it will not be used, otherwise it is either busy if it is currently servicing packets or idle if it is waiting for packets to arrive. A component uses different amounts of electricity in each state. The total energy consumption of a data center is the sum of the energy consumed by all its components:
\begin{equation}
	E_C=\sum_{c\in\mathcal{C}\setminus\mathcal{C}^{\mathsf{vm}}} U_c\cdot E^A+(1-U_c)\cdot E^I,
	\label{eq:energy}
\end{equation}
where $\mathcal{C}^{\mathsf{vm}}$ is the set of VMs and $U_c$ is the active period of the data center component $c$. To calculate $U_c$, we need to consider both single- and multiple-queue devices. The active period of a queue is given by \cite{Kleinrock75}:
\begin{equation}
	\overline{U}_c =
	\begin{cases}
		0,                           & \rm{if} \ \lambda=0    \\
		\frac{1-\rho}{1-\rho^{K+1}}, & \rm{if}\ \lambda\le\mu \\
		\frac{1}{K+1},               & { \rm{otherwise} }
	\end{cases}.
\end{equation}
Physical switches can be modeled with a single-queue for their buffers. Hence the active period of a switch $U_c$ is equal to the active period of its queue:
\begin{equation}
	U_{c_\mathsf{sw} \in\mathcal{C}^{\mathsf{sw}}} = \overline{U}_{c_\mathsf{sw}},
\end{equation}
where $\mathcal{C}^{\mathsf{sw}}$ is the set of switches. A server has multiple buffers: one for the virtual switch and one for each VNF. The server is \texttt{idle} when no packets are being processed at any of its buffers. Thus, the utilization of a server is calculated as:
\begin{equation}
	U_{c_{\mathsf{sr}}\in\mathcal{C}^\mathsf{sr}}=1-\left(1-\overline{U}_{c_\mathsf{sr}^\mathsf{vs}} \right)\cdot\prod_{c_{\mathsf{v}}\in\mathcal{A}_{c_{sr}}}\left(1-\overline{U}_{c_{\mathsf{v}}} \right),
\end{equation}
where $\mathcal{C}^\mathsf{sr}$ is the set of servers and $c_\mathsf{sr}^\mathsf{vs}$ is the virtual switch of the server.

QMs can be distinguished by the assumptions that they make. For the VNFPP, most QMs will assume that traffic arrives according to a Poisson distribution and that they a packet is serviced and leaves the queue according to an exponential distribution. The key distinguishing factor is whether the model assumes bounded or unbounded length queues.

\subsubsection{Unbounded Queues}
Unbounded QMs assume that the queue can grow to be infinitely long. This assumption is unrealistic and exhibits two inaccuracies when used in practice. First, the QM will report that the component packet drop probability $P^d_c = 0$, irregardless of the arrival rate. Second, if the arrival rate exceeds the service rate at any component, the expected length and waiting time of the queue will tend towards infinity. As a result, any service that visits the component will have infinite latency and the solution will be infeasible. In practice, as the arrival rate approaches the service rate, the packet drop probability increases, limiting the maximum length of the queue. Despite these inaccuracies, unbounded QMs are widely used on the VNFPP \cite{GouarebFA18,LeivadeasFLIK18,QuZYSLR20}.

The component utilization, waiting time and packet loss for an unbounded queue can be calculated using standard queueing formula \cite{Kleinrock75}. The component waiting time is given by:
\begin{equation}
	\label{eq:mm1_time_in_component}
    W_c =
    \begin{cases}
        \frac{1}{\mu_c - \lambda}_c, & \text{if } \lambda_c > \mu_c \\
        \infty,                  & \text{otherwise.}
    \end{cases}
\end{equation}
The component utilization is given by:
\begin{equation}
	\label{eq:mm1_utilization}
    U_c = \frac{\lambda_c}{\mu_c},
\end{equation}
and the packet loss $P^d_c = 0$.

\subsubsection{Bounded Queues}
More realistic models acknowledge the presence of packet loss in the network. In the case of bounded queues, each queue has a finite maximum length. If a packet arrives whilst the queue is full, it is dropped and the packet is lost. Hence in a bounded QM, the traffic rate leaving the component will be less than the arrival rate. Additionally, the average queue length grows as the queue utilization increases and so the greater the arrival rate relative to the service rate, the greater the expected packet loss. 

Packet loss can introduce complex interactions between components that are difficult to model. If one component has a high utilization, other components that appear later on the same service path will have a lower arrival rate as a result. If the same component is visited multiple times on the same path, the arrival rate at the component becomes dependent on its own packet loss. The majority of works that use bounded queueing models ignore these interactions and instead calculate the arrival rate at the start of the system before the interactions can occur \cite{ChuaWZSH16,MarottaZDK17}. Recently, we proposed a new method which can be used to accurately calculate the arrival rate at each component \cite{BillingsleyLMMG22}.

Regardless of how the arrival rate is deduced, the component utilization, waiting time and packet loss can be calculated using standard queueing formula for bounded queues \cite{Kleinrock75}. The component waiting time is given by:
\begin{equation}
	W_c=\overline{N}/\hat{\lambda}_c,
    \label{eq:mm1k_wt}
\end{equation}
where $\hat{\lambda}_c$ is the effective arrival rate for the component $c$, $\hat{\lambda}_c = \lambda_c\cdot\left(1-P^d_c \right)$, and $\overline{N}$ is the expected queue length at the component $c$~\cite{Kleinrock75}:
\begin{equation}
	\overline{N} = \begin{cases}
		\frac{\rho[1 - (N^M + 1)\rho^{N^M} + N^M\rho^{N^M+1}]}{(1 - \rho)(1 - \rho^{N^M+1})} , & \text{if } \ \lambda \neq \mu \\
		N^M/2,                                                                      & \text{otherwise.}
	\end{cases}
\end{equation}
where $\rho=\lambda_{c}/\mu_{c}$ and $N^M$ is the maximum queue length. The packet loss $P^d_c$ is given by:
\begin{equation}
    P^d_{c}
	=
	\begin{cases}
		\frac{(1-\rho)\rho^{N^M}}{1-\rho^{N^M+1}}, & \text{if}\ \lambda\neq\mu\\
        \frac{1}{N^M+1}, & \text{otherwise}.
	\end{cases},
	\label{eq:mm1k_pl}
\end{equation}
And the component utilization, $U_c$ is given by:
\begin{equation}
    U_c = \frac{1 - \rho_c}{1 - (\rho_c)^{N^M}}
    \label{eq:mm1k_utilization}
\end{equation}
where $\rho_c = \lambda_c / \mu_c$. 

\subsection{Heuristic models}
Heuristic models do not calculate the objectives directly but instead optimize metrics about the solution which are expected to correlate with the objectives. Whilst they are often used as a fast replacement for accurate models, existing heuristic models do not capture the tradeoffs between the objectives, resulting in a loss of diversity on multi-objective VNFPPs \cite{BillingsleyLMMG22}. In this paper, we propose a new heuristic model which uses the ratio between the arrival rate and service rate, $\rho_c  = \lambda_c / \mu_c$, as a heuristic for the latency and packet loss.

Intuitively, each objective is minimized by reducing the queue length, which depends on the ratio $\rho_c$. The conflict between objectives occurs because adding more VNFs will distribute traffic over more components, reducing the arrival rate, but also increasing the number of components that may be idle. Hence whilst adding additional VNFs will always reduce the latency and packet loss, it may increase the total energy consumption. 

Using this understanding, we propose a heuristic model that uses the average value of $\rho_c$ of components on each path as a heuristic for the latency and packet loss objectives. To calculate the objectives, we first determine the arrival rate using an unbounded queueing model. Although this model is not accurate, the arrival rate at each component in this model will correlate with the true arrival rate. For the heuristic, the first objective is to minimize the average service utilization, where the service utilization for each service $U_s$ is the expected utilization over the paths:
\begin{equation}
	U_s=\sum_{i=1}^{|R^s|} U_{R^s_i} \cdot P_{R^s_i},
\end{equation}
and the service utilization $U_{R^s_i}$ is calculated with:
\begin{equation}
    U_{R^s_i} = \sum_{c\in R^s_i} \rho_c.
\end{equation}
As the energy consumption is a conflicting objective, we also require the optimization algorithm to minimize the total energy consumption using \pref{eq:energy}.

Finally, other heuristic functions have also been proposed in the literature:
\begin{itemize}
    \item\underline{Constant waiting time or packet loss (CWTPL)}: As in~\cite{VizarretaCMMK17} and~\cite{HawiloJS19}, this model assumes that the waiting time at each data center component is a constant. In addition, we also keeps the packet loss probability at each component as a constant. Based on these assumptions, we can evaluate the latency and packet loss for each service and use the accurate bounded model to determine the the energy consumption. All these constitute a three-objective problem that aims to minimize the average latency, packet loss and total energy consumption.

    \item\underline{Resource utilization (RU)}: As in~\cite{GuoWLQA0Y20} and~\cite{QiSW19}, this model assumes that the waiting time is a function of the CPU demand and the CPU capacity of each VM. In addition, the demand is assumed to determine the packet loss probability as well. Based on these assumptions, we evaluate the latency for each service and apply the accurate bounded model to determine the the energy consumption. All these constitute a two-objective problem that aims to minimize the average latency (and by extension the packet loss) and the total energy consumption.

    \item\underline{Path length and used servers (PLUS)}: This model uses the percentage of used servers to measure the energy consumption (e.g.,~\cite{MiottoLCG19,RankothgeLRL17,LiuZDLGZ18}) and the length of routes for each service as a measure of service latency, packet loss or quality (e.g.,~\cite{LuizelliCBG17,AllegKMA17}). All these constitute a two-objective problem that aims to minimize the path length and the number of used servers.
\end{itemize}

\section{Empirical Study}
\label{sec:evaluation}
We seek to answer the following five research questions (RQs) through our experimental evaluation.
\begin{itemize}
    \item\underline{\textbf{RQ1}}: Are our proposed data structures practical for large scale VNFPPs?
    \item\underline{\textbf{RQ2}}: Does our new initialization operator improve upon our previous operator?
    \item\underline{\textbf{RQ3}}: How does our proposed metaheuristic compare against other parallel and synchronous MOEAs?
    \item\underline{\textbf{RQ4}}: Does our proposed heuristic objective function improve on existing objective functions?
    \item\underline{\textbf{RQ5}}: What are the largest problem instances we can consider and what are the current limiting factors?
\end{itemize}

For each research question, we consider three common network topologies: DCell \cite{GuoWTSZL08}, Fat Tree \cite{Al-FaresLV08}, and Leaf-Spine \cite{Cisco19}, illustrated in \pref{fig:topologies}. Where applicable, we calculate the HV of populations of solutions to judge the convergence and diversity of the population. Due to the disparity in the objectives, in each of these tests we first normalize the objective values of all solutions as described in the appendix. The parameter settings for each algorithm are based on the respective authors' recommendations and are also listed in the appendix.

\begin{table}[t]
    \caption{Category and actual number of servers for each network topology}
    \label{tbl:scales}

    \centering
    \begin{tabularx}{\linewidth}{lXXX}
        \toprule
        \multirow{2}{6em}{Number of servers} & \multicolumn{3}{c}{Topology}                      \\
        \cmidrule{2-4}
                                             & Fat Tree                     & Leaf-Spine & DCell \\
        \midrule
        500                                  & 423                          & 512        & 420   \\
        1000                                 & 1024                         & 968        & 930   \\
        2000                                 & 2000                         & 2048       & 1806  \\
        4000                                 & 3456                         & 4050       & 3192  \\
        8000                                 & 8192                         & 7938       & 8190  \\
        16000                                & 16000                        & 15842      & 17556 \\
        32000                                & 35152                        & 31752      & 33306 \\
        64000                                & 65536                        & 64082      & 57840 \\
        \bottomrule
    \end{tabularx}
\end{table}

\subsection{Distance Table Compression Evaluation}
\label{sec:distance_compression}

\begin{figure*}[t!]
    \centering
    \includegraphics[width=\linewidth]{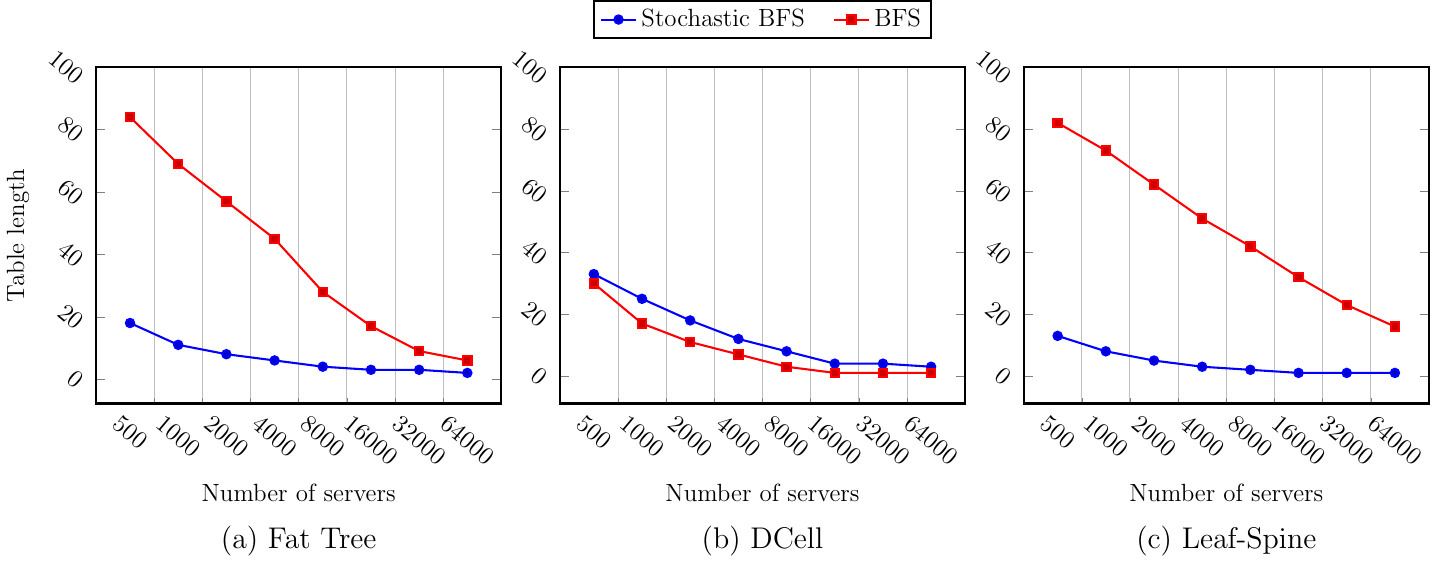}

    \caption{The minimum setting of $N^T$ that was required to reach 99.9\% reliability.}
    \label{fig:distance_compression_alpha}
\end{figure*}

\subsubsection{Methods}
We first aim to determine the minimum practical size of the distance tables. We use monte-carlo methods to find a minimum acceptable size of the distance tables $N^T$. Smaller settings of $N^T$ will grant larger memory savings but also reduce the probability of finding a feasible placement for all VNFs of all services in a solution. We defined an acceptable setting of $N^T$ as one which will place all service instances for $99.9\%$ of a diverse set of solutions where each solution requires at least 90\% data center utilization. Since it is unlikely that a data center will reach this level of utilization, this ensures that the parameter setting is robust against all but the most extreme situations. Similarly, the 99.9\% threshold ensures that on average most initial solutions in a population will be feasible, reducing the risk of a bias towards solutions with low numbers of VNFs.

We considered three topologies at increasingly large scales. Due to the structure of each network topology, it is not possible for all topologies to have the same number of servers at each scale. The exact number of servers for each scale are listed in \pref{tbl:scales}.

For each data center topology and scale, we defined 100 problem instances and generated 1000 solutions for each problem. For both the stochastic BFS and the deterministic BFS, we found the minimum acceptable setting of $N^T$ in the range $0, 1, \cdots, 100$ (i.e. for $99900/100000$ solutions, all service instances must be placed). Note that lower settings of $N^T$ directly correlate with lower memory consumption e.g., $N^T = 0.2$ indicates an 80\% memory saving compared to the naive method.

\subsubsection{Results}
\pref{fig:distance_compression_alpha} shows the minimum acceptable setting of $\alpha$ for each method and for each topology at each scale. These figures show that the memory requirements of the distance tables can be greatly reduced by utilizing a low setting of $N^T$.

Notably, lower settings of $N^T$ can be tolerated on larger graphs. On the smallest topologies, we found that at least 70\% of the information contained in the naive routing tables was not required. On the largest topologies, typically 99\% of the information is not required. Further, we find that across topologies of each scale, the memory requirements remain fairly constant. This suggests that the \textit{relative percentage} of servers in each distance table is less important than the \textit{number} of servers. It is plausible that there exists a constant value for the size of each distance table for an acceptable feasible solution probability. Given the memory requirements of each distance table can be made satisfactorily small with the existing approach, we leave further exploration of this outcome to future work.

Additionally, we found that our proposed stochastic BFS typically allowed for lower settings of $N^T$ than the traditional deterministic BFS. The DCell topology is a special case where the deterministic BFS outperformed the stochastic BFS, however we note that this does not indicate that the deterministic BFS is preferable. Our implementation of the deterministic BFS first explores the servers which reside under a separate switch. Since the DCell topology is symmetrical, this ensures that each server appears the same number of times, provided that a small enough setting of $N^T$ is used. If a larger size were used, or if the deterministic BFS were instead to explore the local switch first, the non-deterministic BFS would be preferable since in the deterministic BFS a small number of servers would be overrepresented. In this way, the deterministic BFS is only preferable for specific instances that can only be guaranteed to occur when the topology is known. Hence it is not appropriate for arbitrary graphs.

In partial response to RQ1, the effectiveness of our proposed distance table compression algorithm enables us to consider far larger problem instances of all topologies.

\subsection{Routing Table Compression Evaluation}
\label{sec:routing_compression}

\subsubsection{Methods}
Without compression, the routing tables required for the mapping process can require impractical amounts of memory. To determine whether compression can reduce the memory requirements to a practical level, we constructed routing tables with and without compression for each topology and at different scales.

\subsubsection{Results}

\begin{figure*}[t!]
    \centering
    \includegraphics[width=\linewidth]{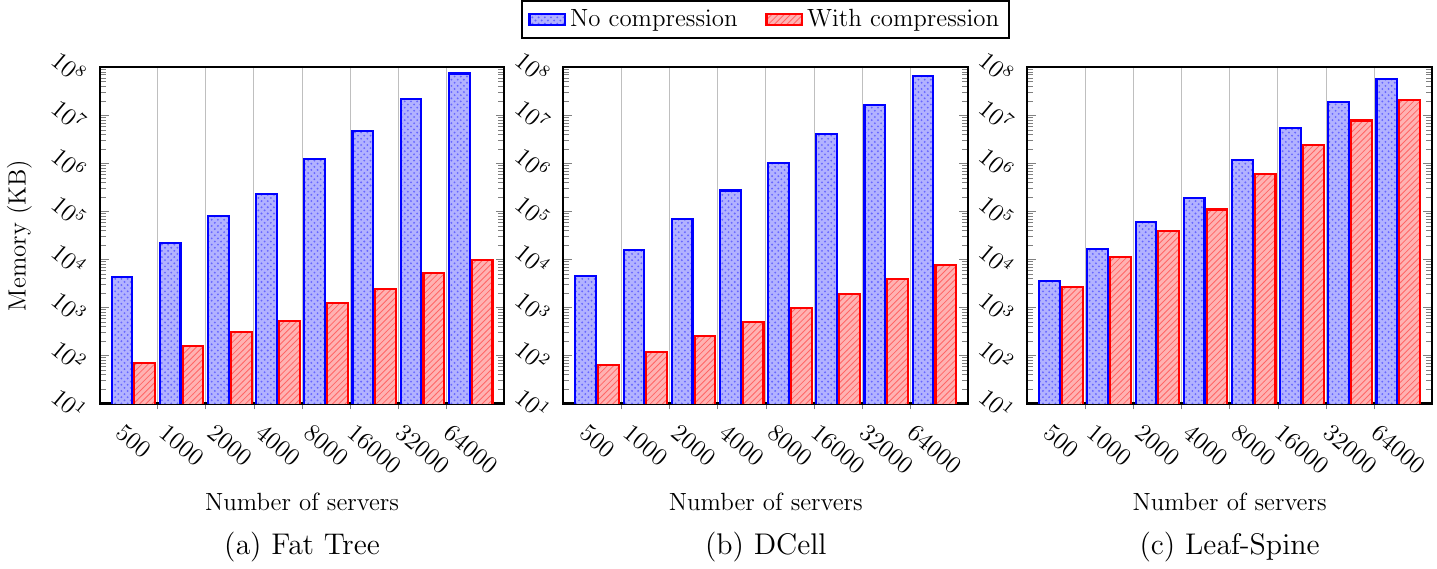}

    \caption{The comparative memory consumption of the routing tables with and without our proposed compression strategy.}
    \label{fig:rt_compression}
\end{figure*}

\begin{table}[t]
	\vspace{1.5em}
    \caption{The percentage of memory saved by using our compressed routing table.}
    \label{tbl:percent_memory}

    \centering
    \begin{tabularx}{\linewidth}{lXXX}
        \toprule
        \multirow{2}{6em}{Number of servers} & \multicolumn{3}{c}{Topology}                      \\
        \cmidrule{2-4}
                                             & Fat Tree                     & Leaf-Spine & DCell \\
        \midrule
        500                                  & 98.38                        & 98.63      & 24.68 \\
        1000                                 & 99.28                        & 99.26      & 31.00 \\
        2000                                 & 99.61                        & 99.65      & 36.53 \\
        4000                                 & 99.77                        & 99.82      & 41.26 \\
        8000                                 & 99.90                        & 99.91      & 48.84 \\
        16000                                & 99.95                        & 99.95      & 54.67 \\
        32000                                & 99.98                        & 99.98      & 59.31 \\
        64000                                & 99.99                        & 99.99      & 63.10 \\
        \bottomrule
    \end{tabularx}
\end{table}

\pref{fig:rt_compression} shows the amount of memory that the routing tables of each topology required with and without compression. The figure shows how the uncompressed routing tables are impractical for large data centers and also that our proposed compression technique uses significantly less memory than the naive approach.

The routing table compression was most effective on hierarchical data centers, e.g. the Fat Tree and Leaf-Spine topologies. We found that hierarchical topologies are particularly amenable to compression. When all servers communicate with other servers via a switch, all paths for a server can be compressed into 2 or 3 rows (the servers with IDs less than the current server, the current server, and the servers with IDs after the current server). Similarly, the hierarchy of the topology ensures that each port of a switch will allow access to either a single server or a cluster of sequential servers which can be compressed into one row. As a consequence, the compressed routing tables used $\sim$3 orders of magnitude less memory than the uncompressed routing tables.

The compressed DCell routing tables also required significantly less memory, specifically 44.92\% less memory on average (see \pref{tbl:percent_memory}), but still required on the order of 10 - 100 GBs of memory to store the larger topologies. This is because of two related factors. First, the DCell topology uses direct connections between servers. As a consequence, each server requires at least one row for each port. Further, the direct server connections mean that although many servers may be accessed via the same component, the IDs of these servers are only contiguous for short stretches, and hence our compression is less effective. Despite these limitations, the compressed DCell topology still uses significantly fewer rows. Second, each row in our compressed formulation requires 50\% more memory. Specifically, the naive implementation must store one ID for the destination server and one ID for the next physical component. The compressed implementation must store an additional ID since it requires two IDs to denote the start and end destination. The combination of these two factors causes the compression to be less effective on the DCell topology.

To conclude our response to RQ1, it is clear that the proposed routing table compression algorithm enables us to consider larger problem instances. However, it is disproportionately more effective on hierarchical topologies.

\subsection{Comparison of Alternative Initialization Operators}

\begin{figure*}[t!]
    \includegraphics[width=\linewidth]{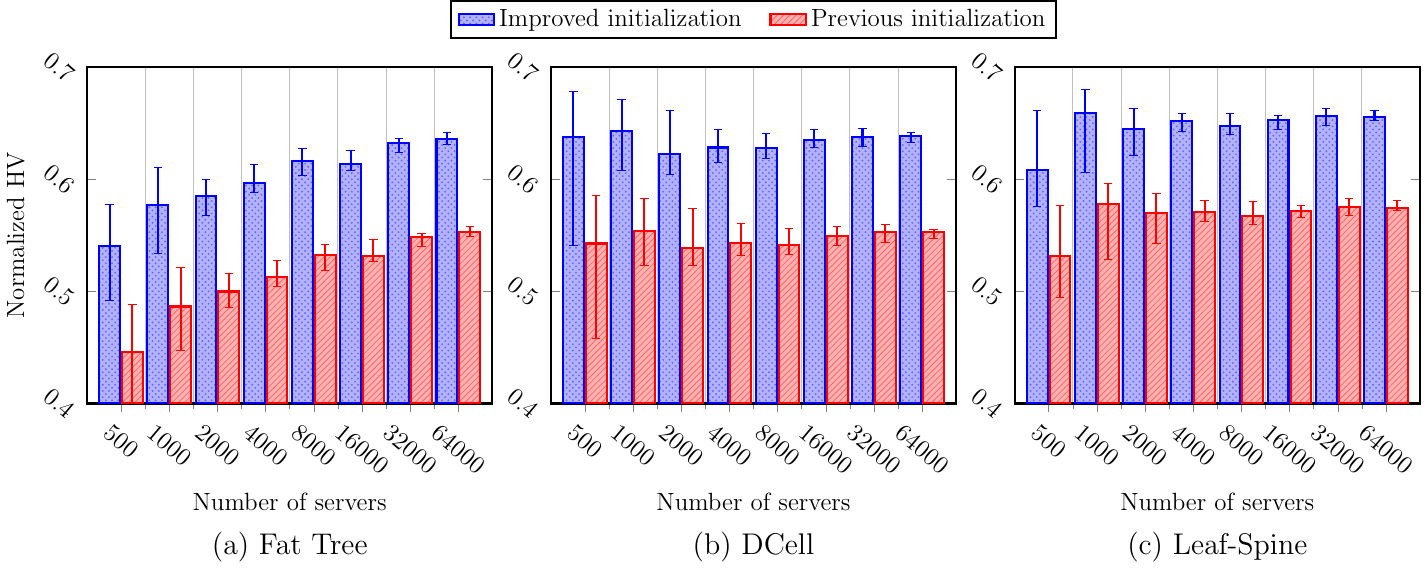}
    \caption{Median and quartile plots of the hypervolume of the initial population using each operator.}
    \label{fig:init_hv}
\end{figure*}

\begin{figure}[t!]
    \includegraphics[width=\linewidth]{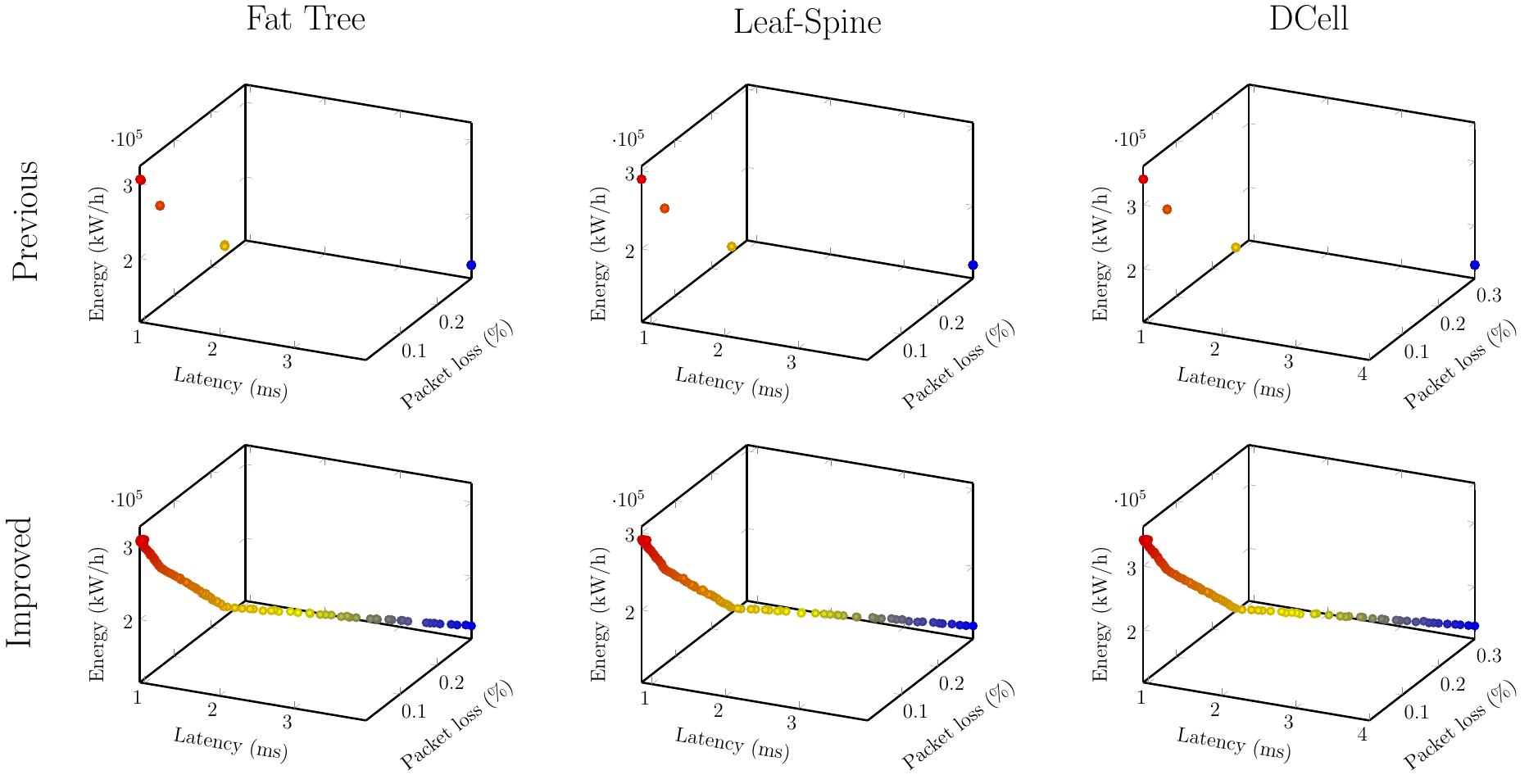}

    \caption{An illustrative example of the non-dominated archives of each initialization operator on a problem instance with 16,000 servers.}
    \label{fig:init_pf}
\end{figure}

\subsubsection{Methods}
To answer RQ2, we compare the effectiveness of our proposed improved initialization operator against an initialization operator we proposed in earlier work \cite{BillingsleyLMMG22}. To evaluate each operator we generated 30 problem instances for each topology and for each data center scale. For each problem we calculated the HV of the initial population.

\subsubsection{Results}
It is clear from \pref{fig:init_hv} and \pref{fig:init_pf}, that our proposed solution produces more evenly distributed individuals than the prior initialization operator. These figures also make it evident that the significantly improved diversity of our improved initialization operator has not come at the cost of solution quality. The prior initialization operator tended to cluster solutions in specific regions since it could only produce solutions where each service instance had the same number of instances. In contrast, our improved initialization operator can vary the numbers of service instances in order to reach a target capacity. This allows for more variance in the energy consumption and average latency and packet loss across solutions.

\subsection{Comparison with Other Approaches}
\subsubsection{Methods}

To answer RQ3, we compare the effectiveness of our proposed algorithm with four state-of-the-art peer MOEAs. Specifically, we compare our algorithm against a similar parallel metaheuristic, PPLS/D \cite{ShiZS20}, and three important MOEA algorithms (NSGA-II \cite{DebAPM02}, IBEA \cite{ZitzlerK04}, MOEA/D \cite{ZhangL07}). We believe this provides sufficient proof-of-concept evidence to extend to otherMOEA variants (e.g.,~\cite{LiZZL09,LiZLZL09,CaoWKL11,LiKWCR12,LiKCLZS12,LiKWTM13,LiK14,CaoKWL14,LiFKZ14,LiZKLW14,WuKZLWL15,LiKZD15,LiKD15,LiDZ15,LiDZZ17,WuKJLZ17,WuLKZZ17,LiDY18,ChenLY18,ChenLBY18,LiCFY19,WuLKZZ19,LiCSY19,Li19,GaoNL19,LiXT19,ZouJYZZL19,LiuLC20,LiXCT20,LiLDMY20,WuLKZ20,BillingsleyLMMG19,LiX0WT20,WangYLK21,LiLLY25,WilliamsLM25,LiYV25,ChenDTQL25,XuLLJL25,ZhouHS024,ChenDLTC24,XuLL24,ZouSLHYZL24,ZhangDFTL24,LiCY24,LiGX23,XuLA22,WilliamsLM23,Williams023,ChenL23,WilliamsLM232,HuangL23,WilliamsL23,LiC23,LiNGY22,ChenLTL22,LiLLM21}).

\begin{figure*}[t!]
    \includegraphics[width=\textwidth]{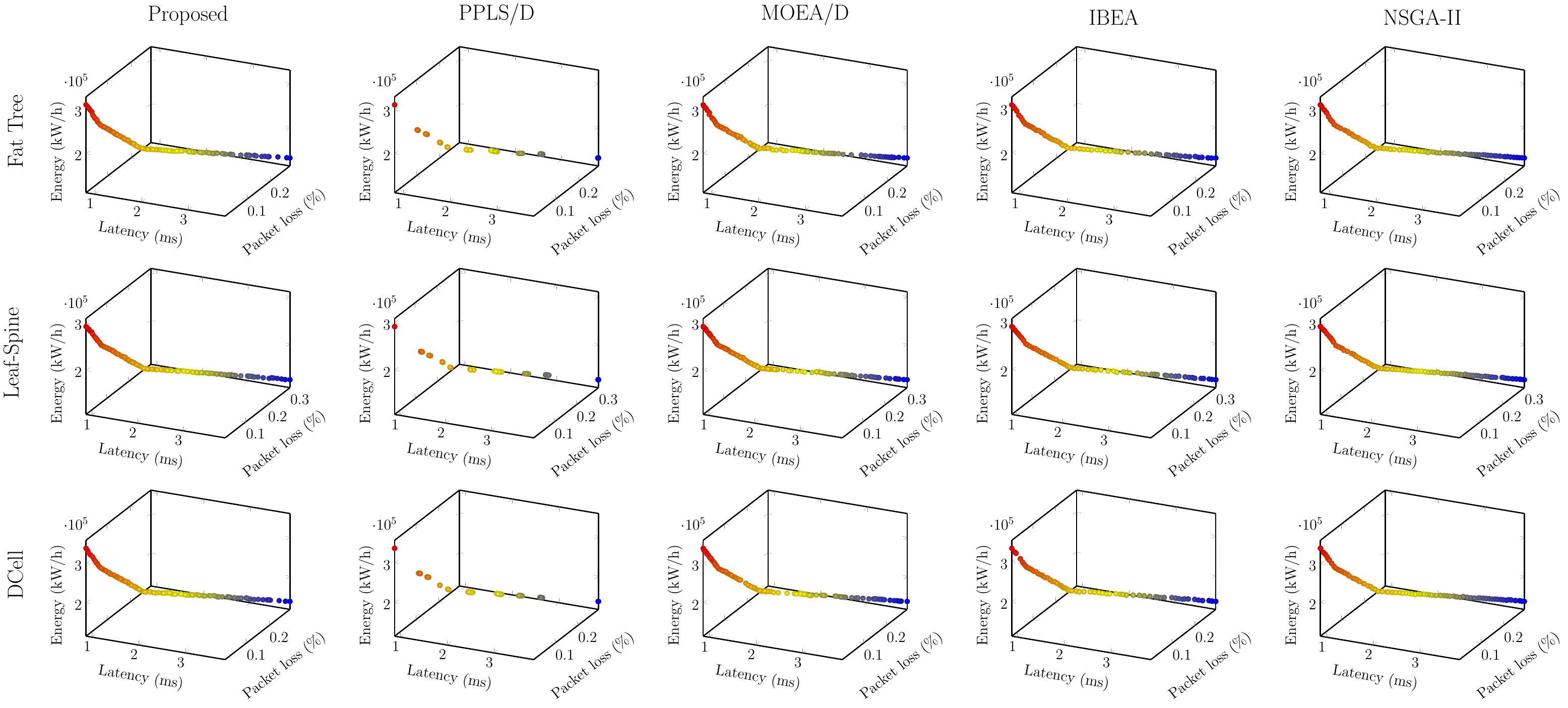}
    \caption{An illustrative example of the non-dominated archives of each algorithm on a problem instance with 16,000 servers.}
    \label{fig:alg_pf}
\end{figure*}

\begin{table}[t!]
	\vspace{2em}
    \caption{Median wall clock time (s) as a multiple of the median wall clock time of our proposed algorithm averaged over all topologies.}
    \label{tbl:time_multiple}
    \centering
    \resizebox{.6\columnwidth}{!}{%
        \begin{tabular}{lllllll}
            \toprule
            \multirow{2}{6em}{Algorithm} & \multicolumn{6}{c}{Number of servers}                                     \\
            \cmidrule{2-7}
                                         & 500                                   & 1000 & 2000 & 4000 & 8000 & 16000 \\
            \midrule
            NSGA-II                      & 5.00                                  & 4.38 & 4.05 & 3.80 & 3.32 & 3.13  \\
            IBEA                         & 4.44                                  & 4.03 & 3.82 & 3.59 & 3.19 & 3.01  \\
            MOEA/D                       & 6.81                                  & 6.08 & 5.65 & 5.35 & 4.71 & 4.44  \\
            PPLS/D                       & 0.94                                  & 0.95 & 0.94 & 0.96 & 0.97 & 0.99  \\
            \bottomrule
        \end{tabular}%
    }
\end{table}

\begin{figure}[t!]
    \centering
    \begin{subfigure}[b]{.7\linewidth}
        \includegraphics[width=\textwidth]{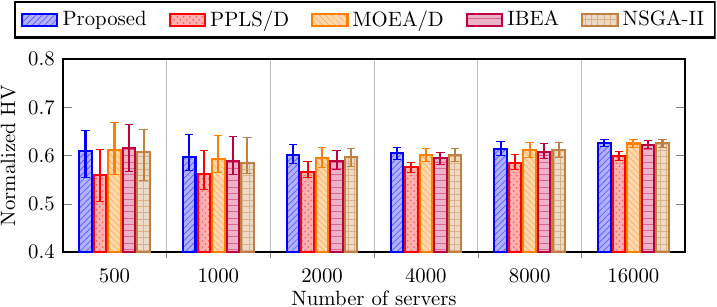}
        \caption{Fat Tree}
    \end{subfigure}\vspace{1em}
    
    \begin{subfigure}[b]{.7\linewidth}
        \includegraphics[width=\textwidth]{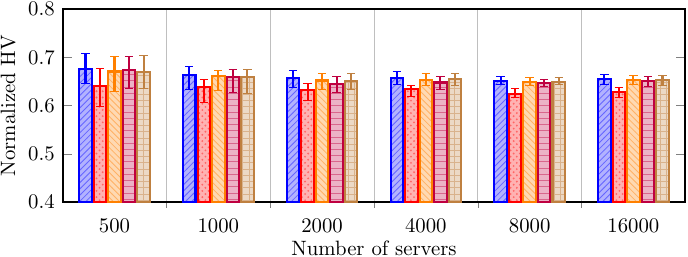}
        \caption{Leaf-Spine}
    \end{subfigure}\vspace{1em}
    
    \begin{subfigure}[b]{.7\linewidth}
        \includegraphics[width=\textwidth]{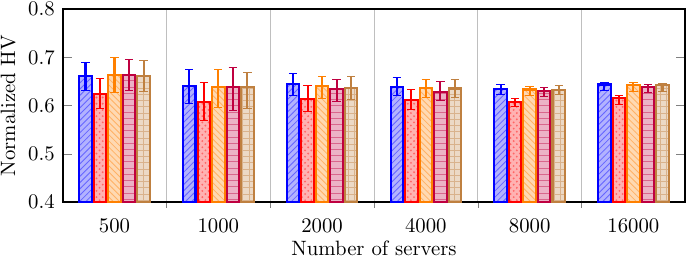}
        \caption{DCell}
    \end{subfigure}

    \caption{Median and quartile plots of the hypervolume of each algorithm.}
    \label{fig:alg_hv}
\end{figure}

PPLS/D is a decomposition algorithm which uses local search to solve each subproblem in parallel. However, unlike our proposed algorithm, PPLS/D requires one thread per subproblem and hence its effectiveness as an optimization algorithm is dependent on the hardware it is executed on. PPLS/D maintains a population of unexplored solutions and an archive of non-dominated solutions for each subproblem. First, an initial solution is added to the unexplored population of each subproblem. Next, for each subproblem the algorithm selects the best solution from the unexplored population and generates a set of neighboring solutions. A neighboring solution is added to the unexplored population and the non-dominated archive if it is: 1) closer to the subproblem that generated it than any other subproblem, as measured by the angle between the solution and the subproblems and 2) is either a better solution to the subproblem than the current best solution or if it is not dominated by any solutions in the non-dominated set of the subproblem. Once the stopping condition is met, the algorithm aggregates the non-dominated archives from each subproblem and returns the overall non-dominated solutions.

PPLS/D was designed for unconstrained optimization problems, so require modification to fit the VNFPP. In our variant, an infeasible solution is accepted into the unexplored population only if there is not a more feasible solutions in the population. The original implementation of PPLS/D also explores all neighbors of each solution in the final population. Given the very large neighborhoods present in the VNFPP, this step was removed.

To evaluate each algorithm we generated 30 problem instances for each topology and for several data center scales. Since NSGA-II and IBEA do not utilize an external population, they can only discover a fixed number of solutions. To ensure that the other algorithms do not have an unfair advantage, we find a subset of solutions from each archive the same size as the population size of IBEA and NSGA-II. A common method of reducing the size of a non-dominated archive is to use a clustering algorithm and then to select a representative solution for each cluster \cite{ZioB11}. In this work we use $k$-means clustering \cite{HartiganW79} to generate the same number of clusters as the population size of IBEA and NSGA-II, and then select the nearest non-dominated solution to the centroid of each cluster from the archive population.

For each problem we recorded the wall-clock execution time and the HV of the final population or subset. We allowed each algorithm to run until they had performed 12,000 evaluations. All tests were run on a 8 core/16 thread CPU at 2.6GHz. All parallel algorithms were implemented using the parallelization library Rayon\footnote{https://github.com/rayon-rs/rayon}.



\subsubsection{Results}
A key benefit of parallelizing is its potential to reduce the execution time. \pref{tbl:time_multiple} shows that the parallel algorithms require multiple times less execution time than comparable sequential metaheuristics. Specifically, our algorithm has a median wall clock time between 3.01 - 6.81$\times$ faster than IBEA, NSGA-II and MOEA/D. In this regard, our proposed algorithm significantly out performs state of the art, serial MOEAs.

Further, \pref{fig:alg_hv} shows that the fast execution time of our proposed algorithm does not come at the cost of solution quality. From \pref{fig:alg_pf}, we can see that all algorithms tended to find similar populations of solutions. In this regard our proposed algorithm performs as well as other state of the art MOEAs on the VNFPP. Critically, the high quality solutions of most algorithms appear to be a consequence of the initialization operator. It is clear from \pref{fig:init_pf} and \pref{fig:alg_hv} that the metaheuristics were only able to make small improvements over the solutions found by the improved initialization operator. As a result, any algorithm that maintains a diverse population of solutions will be fairly effective.

In contrast, PPLS/D suffers from two issues related to population diversity. First, the algorithm has specific rules about which solutions can be included in the total archive which causes it to disregard much of the initial population. Second, we can see from \pref{fig:alg_pf} that the algorithm is not effectively exploring solutions away from the subproblems and hence cannot recover the information it has lost. This hints at a potential drawback of the local search approach: since the neighborhood of each solution is very large, it is only possible to search a small region around the initial solution using local search. Since the number of subproblems in PPLS/D is limited by the number of available threads, PPLS/D only explores a small region of the search space and hence represents a small region of the Pareto front. As our algorithm considers initial solutions in the initial archive and, as is it is not limited by specific properties of the hardware it is executed on, it does not encounter these issues.

Overall, it is clear that our algorithm is more efficient than the alternative serial MOEAs considered, and ensures higher quality solutions than a comparable parallel MOEA. However, it is not clear that the philosophy behind our proposed algorithm (i.e. favouring local search over global search) is beneficial for the VNFPP.

\subsection{Efficient Alternative Models}
\subsubsection{Methods}
An efficient and effective model that aligns with our objective functions is critical for the overall performance of our algorithm. To answer RQ4, we compare the performance of our proposed utilization model against each of the other objective functions introduced in \pref{sec:practical_objective_functions} (i.e. accurate, M/M/1/K, M/M/1, PLUS, RU, and CNST). Due to the efficiency and effectiveness of our proposed algorithm, we use it as the optimization algorithm for each test.

To evaluate each model we generated 30 problem instances for each topology and for several data center scales. We used our proposed algorithm to solve each problem for each model. We allowed the algorithm to run until it had performed 12,000 evaluations. For each problem we recorded the wall-clock execution time and calculated the HV of the final population.

\subsubsection{Results}
\pref{fig:model_hv} shows that our proposed heuristic under-performs in comparison to both accurate models in terms of the HV, but it significantly outperforms all alternative heuristics. This indicates that our heuristic captures \textit{most} of the information on the tradeoffs between objectives. The discrepancy is likely because the utilization of a component changes \textit{linearly} with changes to the arrival rate, whilst the latency and packet loss change \textit{non-linearly}. As a result, the utilization heuristic would consider two possible paths with the same sum utilization as equal, even if one path utilizes an overloaded component with a very high arrival rate and hence high latency and packet loss. The larger discrepancy on small problems reflects the optimization algorithms ability to identify and rectify this situation easier when there are fewer variables to consider.



Despite this, \pref{tbl:model_time_multiple} shows that the utilization heuristic is typically orders of magnitude faster than both accurate models and also outperforms some existing heuristic approaches. Specifically, the mean wall clock execution time of our heuristic is 53.33 - 4.37$\times$ faster than the accurate model and 20 - 2.23$\times$ faster than the M/M/1/K model. Notably, these results are significantly affected by the DCell topology where the execution times are closer.

The superior performance of our proposed heuristic in terms of execution time is due to multiple factors. Principally, the heuristic leads to fewer operations overall. Clearly, the heuristic requires fewer operations than more accurate models which will directly lead to performance improvements. Indirectly, since the heuristic forms a two objective problem, it is faster to determine whether one solution dominates another. Further, since the objective space is smaller, the non-dominated archive will contain fewer solutions on average, accelerating the process of calculating the non-dominated set.

\begin{figure}[t!]
	\centering
    \includegraphics[width=\linewidth]{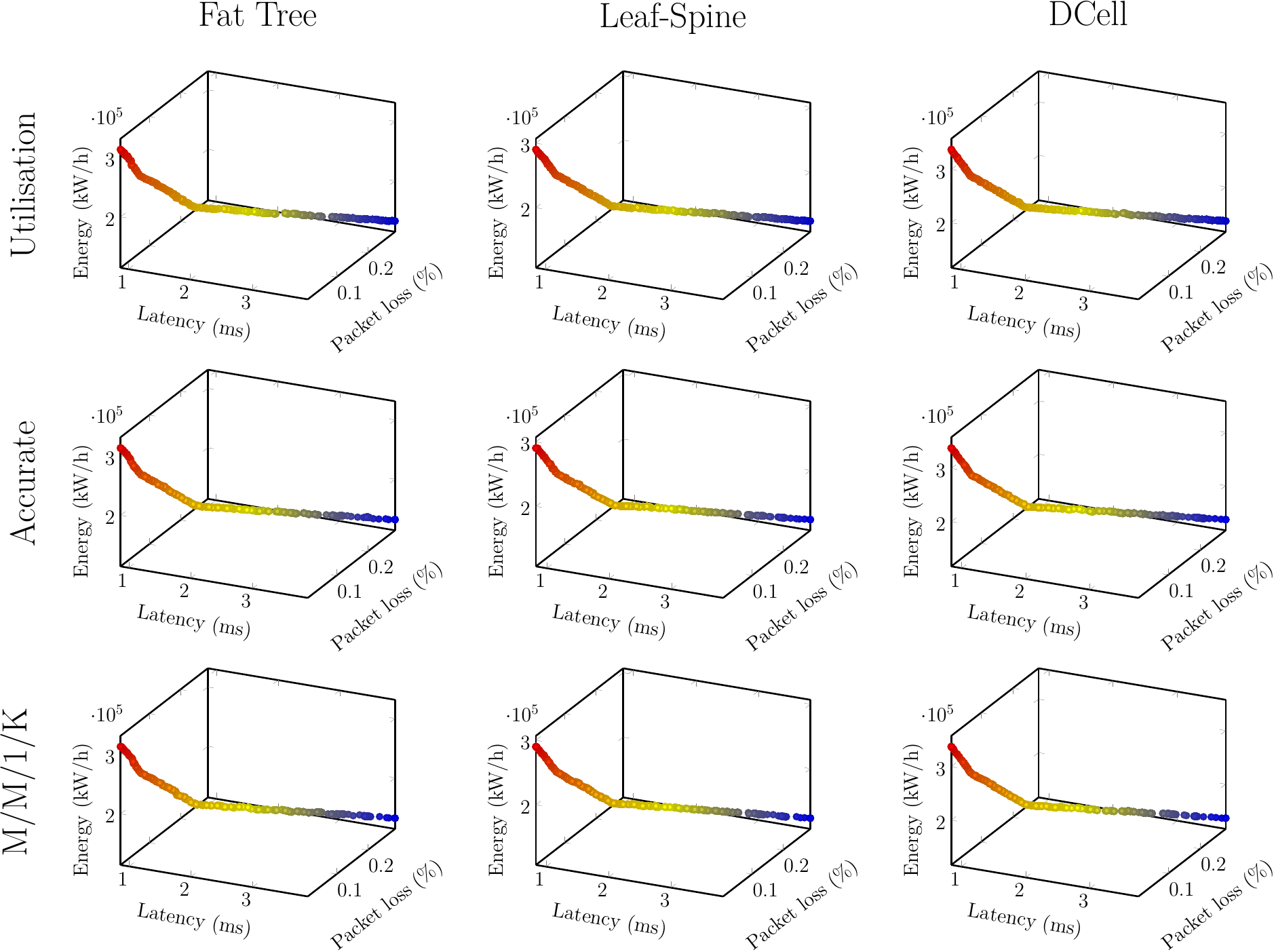}
    \caption{An illustrative example of the non-dominated solutions found by our proposed heuristic and both accurate models for problems with 16,000 servers.}
    \label{fig:models_pfs}
\end{figure}
\begin{figure}[t]
    \centering
    \begin{subfigure}[b]{0.9\linewidth}
    	\centering
        \includegraphics[width=.9\textwidth]{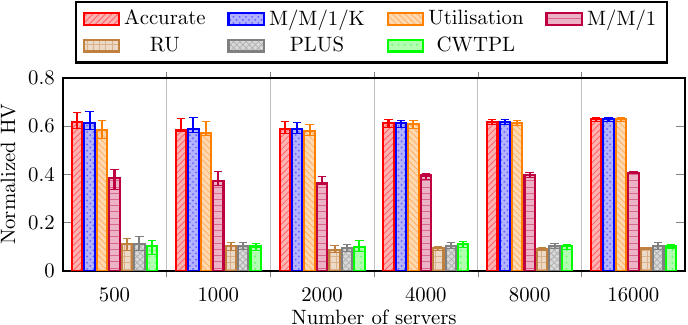}
        \caption{Fat Tree}
    \end{subfigure}
    
    \begin{subfigure}[b]{0.9\linewidth}
    	\centering
        \includegraphics[width=.9\textwidth]{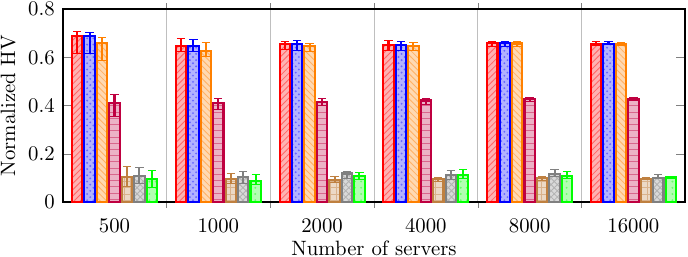}
        \caption{Leaf-Spine}
    \end{subfigure}
    
    \begin{subfigure}[b]{0.9\linewidth}
    	\centering
        \includegraphics[width=.9\textwidth]{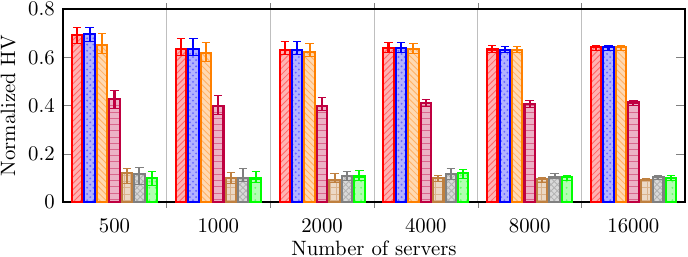}
        \caption{DCell}
    \end{subfigure}

    \caption{The median and quartiles of the normalized HV using our proposed algorithm and each model.}
    \label{fig:model_hv}
\end{figure}

\begin{table}
    \caption{Median wall clock time (s) as a multiple of the median wall clock time of our proposed model averaged over all topologies.}
    \label{tbl:model_time_multiple}

    \centering

    \resizebox{.5\columnwidth}{!}{%
        \begin{tabular}{lllllll}
            \toprule
            \multirow{2}{4em}{Model} & \multicolumn{6}{c}{Number of servers}                                       \\
            \cmidrule{2-7}
                                     & 500                                   & 1000  & 2000  & 4000 & 8000 & 16000 \\
            \midrule
            Accurate                 & 53.33                                 & 13.33 & 12.50 & 8.88 & 6.08 & 4.37  \\
            M/M/1/K                  & 20.00                                 & 5.33  & 5.07  & 3.80 & 2.80 & 2.23  \\
            M/M/1                    & 1.00                                  & 0.40  & 0.57  & 0.65 & 0.59 & 0.52  \\
            RU                       & 10.00                                 & 2.00  & 2.00  & 1.45 & 1.04 & 0.79  \\
            PLUS                     & 1.00                                  & 0.10  & 0.04  & 0.20 & 0.21 & 0.25  \\
            CWTPL                    & 10.00                                 & 2.67  & 2.43  & 1.60 & 1.13 & 0.85  \\
            \bottomrule
        \end{tabular}%
    }
\end{table}

Overall, the utilization heuristic presents a viable trade off between wall clock execution time and solution quality. Whilst some of the other heuristic approaches require less time still, the impact on solution quality and diversity is too severe. The proposed heuristic is likely better suited for larger problem instances where the execution time of more accurate models could be prohibitively time consuming.

\subsection{Applications to Large Data Centers}
\subsubsection{Methods}
Finally, to fully understand the scalability of our algorithm, we combined the most effective components as measured in previous tests and applied the resultant algorithm to increasingly large problem instances. Specifically, we integrated the utilization model, with our proposed operators and compression techniques into our proposed algorithm. Our aim is to identify which, if any, of these components prevent the algorithm from being applied to arbitrarily large VNFPPs.

We generated 30 problem instances for each topology and for increasingly large data center scales. Since solving a large problem instance can become time consuming, each data center scale contains approximately twice as many servers as the previous scale (i.e. $500$, $1000$, $2000$, $\dots$). We allowed the algorithm to run until it had performed 12,000 evaluations. We terminated testing when it was no longer practical to solve problem instances. For each problem we recorded the wall-clock execution time and calculated the HV of the final population.


\begin{figure*}[t]
	\centering
    \includegraphics[width=\linewidth]{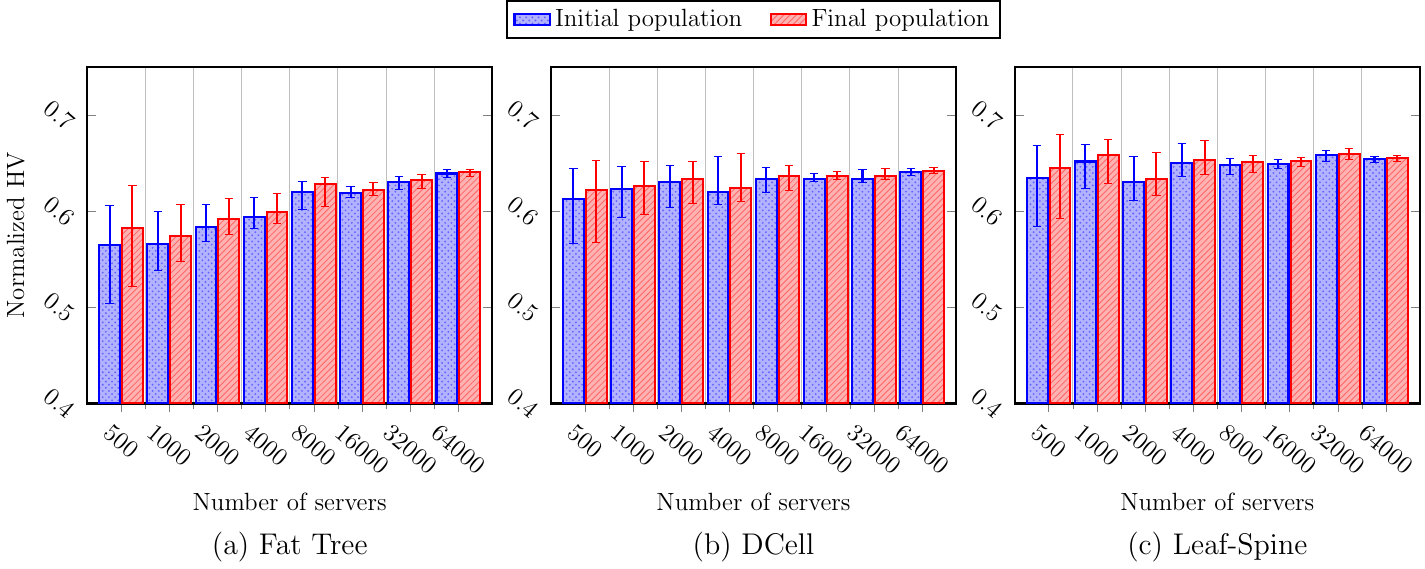}
    \caption{Hypervolume values for each topology and scale}
    \label{fig:large_hv}
\end{figure*}

\begin{figure}[t]
	\centering
    \includegraphics[width=.8\linewidth]{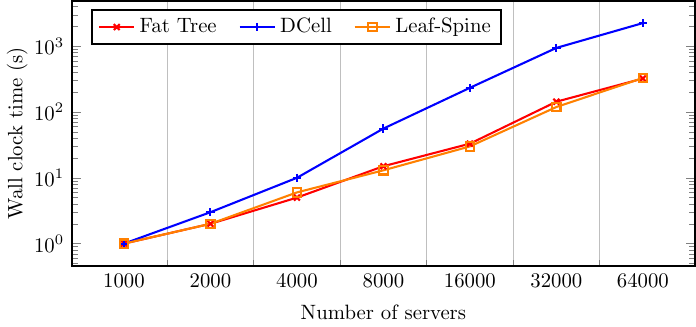}
    \caption{Wall clock execution time for each topology and scale}
    \label{fig:large_times}
\end{figure}

\subsubsection{Results}
\pref{fig:large_hv} demonstrates the high scalability of our algorithm. Specifically, our proposed algorithm can solve problems with up to 64,000 servers. We found that the routing table construction step prevented our algorithm from scaling to larger scale problems. The BFS based routing table construction algorithm has a low theoretical quadratic time complexity that scales with the number of components. In practice, a quadratic time complexity is unsuitable for very large topologies with hundreds of thousands of physical components. By way of illustration, we applied the routing construction algorithm to the first 1/10th of the servers in a DCell network topology with $\sim$128,000 servers. The construction of these routing tables took $\sim$124 hours ($\sim$5 days). By extrapolation, we can calculate the full construction of the routing tables would require at least 7 weeks to complete. Notably, unless otherwise resolved, this limitation applies to other optimization algorithms, including heuristics, that place VNFs in arbitrary networks.

As \pref{fig:large_times} shows, the other components of the algorithm demonstrated promising results on large scale topologies. Once preprocessing was complete, the algorithm execution time required on the order of seconds to minutes for 500-32000 servers. The largest data center topologies had a median time complexity of between $\sim$27 minutes (Leaf-Spine, 1631 seconds) and $\sim$70 minutes (DCell, 4225 seconds). 

It is interesting to note that the average execution time of the algorithm is significantly higher on the DCell topology. This could be explained by differences in the availability of cached resources. Whilst the data center is small, a larger proportion of the required memory can be held in a cache. However, on larger problems, \textit{cache prediction} will become more important. Cache prediction algorithms aim to retrieve data before it is requested. One assumption in cache prediction is the principle of spatial locality: the assumption that future memory accesses are likely to occur near to recent ones \cite{KennedyM92}. Since the DCell network topology allows servers with distant IDs to communicate, this assumption will be violated, and memory accesses are less likely to use a cache. In contrast, on the Fat Tree and Leaf-Spine topologies, nearby servers have similar IDs and hence are stored nearby in memory.

\pref{fig:large_hv} also shows that the initialization operator produces populations that are competitive with those found by our metaheuristic and hence could function as a standalone heuristic. It is clear from the results of earlier tests that the initialization operator is very effective as it is considers both the number of instances (through the improved initialization operator) and their placement (through our genotype-phenotype solution representation). Overall, our metaheuristic produces populations with a higher median HV, but the overall results are not significantly better than the initial population ($p < 0.05$) in any test case. Further, the initialization operator is a simpler algorithm, has a low time complexity and does not need to evaluate solutions. As a result, it is significantly faster on all problem instances. Despite this, since the initialization operator relies on the genotype-phenotype solution representation, it cannot scale to solve larger problem instances than our proposed algorithm.

The good relative performance of the initialization operator can be partly explained by the difficulty of very large problem instances. The largest problem instances in our tests contain on the order of 6000 services. Since the objective functions calculate the mean QoS across all services, a visible improvement to the objective functions requires improving the QoS of many services. However, the optimization algorithm is permitted only 12,000 evaluations to ensure it finishes in an acceptable time. As a result there is little opportunity to find improving solutions for many services. Irregardless, the effectiveness of the initialization operator should not be overlooked as it presents a viable alternative to more complex metaheuristic or exact approaches on very large problem instances.

\section{Conclusions}
\label{sec:conclusion}
In this paper we proposed a new MOEA for arbitrary graphs capable of efficiently solving large scale instances of the VNFPP. To this end we proposed:
\begin{itemize}
    \item A novel solution representation for large scale, arbitrary network topologies.
    \item A novel local search based, parallel MOEA that significantly reduces the total algorithm execution time whilst achieving greater solution diversity.
    \item An improved initialization algorithm that ensures high quality, diverse solutions.
    \item A novel heuristic model that captures sufficient information about the VNFPP whilst greatly reducing the evaluation time.
\end{itemize}
We evaluated the effectiveness of each of these components and found they made significant improvements over state-of-the-art alternatives. By synthesizing these components we demonstrated that our algorithm could solve problem instances with up to 64,000 servers, a 16$\times$ improvement over problems that have been considered in the literature so far. Further, we found that the improved initialization operator combined with the genotype-phenotype solution representation is highly effective on very large problem instances.

Further extensions could be considered in future work. 
\begin{itemize}
    \item Our proposed operators combined produce an effective heuristic for the multi-objective VNFPP. It would be interesting to apply the same principles to the single-objective VNFPP which aim to minimize a cost metric and treat QoS requirements as a constraint.
    \item A challenging problem that remains unresolved is \textit{service resilience} i.e. the capacity for the data center to continue to provide a service when components may fail. This is an instance of a well known NP-Hard problem, graph partitioning, hence approximation algorithms and heuristics will likely be required.
    \item Finally, future work could extend our metaheuristic into a dynamic optimization problem such as in \cite{OtokuraLKKSM16}. Metaheuristics are well suited to dynamic optimization problems and many algorithmic frameworks have been proposed \cite{AlbaNS13}. This would enable our algorithm to adapt to changing requirements and to present the best set of options at any given moment.
\end{itemize}

\section*{Acknowledgment}
K. Li was supported by UKRI Future Leaders Fellowship (Grant No. MR/S017062/1) and Royal Society (Grant No. IEC/NSFC/170243).

\bibliographystyle{IEEEtran}
\bibliography{IEEEabrv,bibliography,knee}

\end{document}